\pdfoutput=1
\documentclass[journal]{IEEEtran}
\usepackage[utf8]{inputenc}

\usepackage{adjustbox}
\usepackage{graphicx}

\usepackage{amsmath}
\usepackage{amssymb}
\usepackage{amsthm}

\usepackage{booktabs}
\usepackage{longtable}
\usepackage{array}
\usepackage{multirow}
\usepackage{wrapfig}
\usepackage{float}
\usepackage{colortbl}
\usepackage{pdflscape}
\usepackage{tabu}
\usepackage{threeparttable}
\usepackage{threeparttablex}
\usepackage[normalem]{ulem}
\usepackage{makecell}
\usepackage[table,xcdraw]{xcolor}

\usepackage{caption}

\usepackage[linesnumbered,ruled,vlined]{algorithm2e}

\newcommand{\norm}[1]{\left\lVert#1\right\rVert}

\theoremstyle{definition}
\newtheorem{definition}{Definition}

\newtheorem{proposition}{Proposition}

\usepackage{float}

\usepackage{hyperref}

\title{An exact counterfactual-example-based approach to tree-ensemble models interpretability}
\author{
    \IEEEauthorblockN{Pierre Blanchart\IEEEauthorrefmark{1}}
    \\\IEEEauthorblockA{\IEEEauthorrefmark{1}CEA, LIST, 91191 Gif-sur-Yvette Cedex, France
    \\\{pierre.blanchart\}@cea.fr}
}

\hypersetup{
pdfauthor={Pierre BLANCHART}, 
pdftitle={An exact counterfactual-example-based approach to tree-ensemble models interpretability, decision explanation},
pdfkeywords={XGBoost, LightGBM, tree ensemble models, interpretability, counterfactual example, fault diagnosis}
}

\begin{document}

\maketitle

\begin{abstract}
Explaining the decisions of complex machine learning models is becoming a necessity in many areas where trust in ML models decision is essential to their accreditation / adoption by the experts of the domain. The ability to explain models decisions comes also with an added value since it allows to provide diagnosis in addition to the model decision. This is highly valuable in such scenarios as fault/abnormality detection.
Unfortunately, high-performance models do not exhibit the necessary transparency to make their decisions fully understandable. And the black-boxes approaches, which are used to explain such model decisions, suffer from a lack of accuracy in tracing back the exact cause of a model decision regarding a given input. Indeed, they do not have the ability to explicitly describe the decision regions of the model around that input, which would be necessary to exactly say what influences the model towards one decision or the other.
We thus asked ourselves the question: is there a category of high-performance models among the ones commonly used for which we could explicitly and exactly characterise the decision regions in the input feature space using a geometrical characterisation? Surprisingly we came out with a positive answer for any model that enters the category of tree ensemble models, which encompasses a wide range of models dedicated to massive heterogeneous industrial data processing such as XGBoost, Catboost, Lightgbm, random forests... For these models, we could derive an exact geometrical characterisation of the decision regions under the form of a collection of multidimensional intervals.
This characterisation makes it straightforward to compute the optimal counterfactual (CF) example associated with a query point, as well as the geometrical characterisation of the entire decision region of the model containing the optimal CF example. We also demonstrate other possibilities of the approach such as computing the CF example based only on a subset of features, and fixing the values of variables on which the user has no control. This allows in general to obtain more plausible explanations by integrating some prior knowledge about the problem. 

A straightforward adaptation of the method to counterfactual reasoning on regression problems is also envisaged.
\end{abstract}

\section*{Acknowledgement}
This research was partly funded by the Artificial Intelligence for Digital Automation (AIDA) collaborative program. The author also would like to thank Eiji Kawasaki (CEA, LIST) for constructive criticism and proofreading of the manuscript.

\section{Introduction and context}

Tree ensemble models are a very popular class of models which have shown great performance and robustness on heterogeneous tabular data, while requiring very little pre-processing work on the initial data before training the model. They are generally much faster to train than neural networks and do not require any specific hardware. They can also handle a big amount of missing data, and, allow an easier control of overfitting in the case of highly imbalanced datasets. As such, they are widely used to perform fault/anomaly detection on complex industrial datasets which naturally yield imbalanced classes, faults having a much lower occurrence than normal data. The fine-grained monitoring of machines and products in complex industrial manufacturing processes such as production lines, or large multi-instances dynamic systems such as electricity/water transportation networks has created a high demand for this type of models.
 

Fault/anomaly detection is often addressed by learning a global fault detection model (FDM) taking all sensor measurements into account. Among tree ensemble models, gradient boosted trees have prove superior performance on massive heterogeneous industrial data. In particular, we can mention XGBoost~\cite{chen_xgboost:_2016}, a gradient boosting tree ensemble classification/regression method, and a many time fault detection Kaggle challenge winner\footnote{https://www.kaggle.com/c/bosch-production-line-performance/}\footnote{https://www.kaggle.com/c/ieee-fraud-detection/}. LightGBM \cite{ke2017lightgbm} has been currently the most often used implementation of gradient boosted trees due to its better training efficiency, and has also been the model implemented in many Kaggle challenges winning solutions\footnote{https://www.kaggle.com/c/vsb-power-line-fault-detection/}\footnote{https://www.kaggle.com/c/talkingdata-adtracking-fraud-detection/}. This model exhibits a fantastic robustness on highly unbalanced two-class classification problems such as fault detection.

The drawback of such models is that they are not transparent, and do not provide any direct explanation of their decision, which is yet the main component for fault identification and diagnosis~\cite{ribeiro2016should,lipton2016mythos}. Some approaches cope with this issue by simplifying the learned FDM to make it interpretable~\cite{hara2016making,gallego2016interpreting}, but degrading the detection performances. In a similar spirit, some models are constrained to be simple enough to stay transparent as to their decision making~\cite{letham2015interpretable}, impacting the detection performance as well. 

In this work, we focus on explaining individual model predictions, leaving aside the approaches that look for global model interpretability.

Among the methods seeking to do so, counterfactual (CF) methods have grown more and more popular in the literature due to their proximity to the human reasoning \cite{wachter2017counterfactual, mittelstadt2019explaining, keane2021if}, and their resemblance to the everyday life explanations in human conversations. In particular, CF explanations are particularly useful to help people understand how they can change inputs under their control in order to achieve a different outcome (for instance, what they could change in their saving/purchasing habits so that a credit is no longer refused to them).
For classification problems, CF approaches answer the general question: what would need to be changed a minima in the initial data so that it be classified in an other user-defined class.
As such, CF methods can be applied to explain the decision of any problem which can be re-qualified as an anomaly detection model. This encompasses fault detection, and all problems where the purpose is to detect a deviation from a normal state: diseases, fraud, ambiguous miss-classified data ... An application to credit denial, and to ambiguous digit images miss-classification is showcased in the experiments (sec.~\ref{sec:experiments}).

For regression, CF approaches answer the question: what would have to be changed a minima in the initial data, so that the prediction of the model be in a different interval. We show that our approach can be used to to so, as long as it is possible to faithfully learn the relation between the exogenous variables and the target values for the problem at hand using a tree ensemble model.
An application to house sale prices is showcased. In particular, we try to answer the question: what are the minimal changes to perform among a list of predefined possible changes, in order to increase the sale value of a house.

In the next section, we review the interpretability methods which focus on explaining individual predictions, and which are similar in spirit to CF approaches, i.e. try to explain why a data point was classified in a given class by comparing it with a data point from an other class. We then review both model agnostic and model specific CF approaches which have been proposed until now in the literature for tree ensemble model interpretability. In particular, we try to identify the closest work among model specific CF approaches for tree ensemble models, which is the exact categorisation of our approach.

Among other advantages of our method, we can also mention its ability to find the closest CF example in the setting were only a few selected features are allowed to vary, which is not possible with the model specific approaches mentioned above. 

\section{Related work}



\subsection{Non-CF interpretability methods}

\subsubsection{Model agnostic methods}
A common appraoch to model agnostic interpretability is the use of local or global surrogates, i.e. local or global approximations of the initial model with a simpler interpretable model such as decision trees, linear models, or generalized additive models ... A more in-depth description of such approaches can be found for instance in \cite{molnar2019}.
These method suffer from stability problems and/or from the ability of the surrogate to correctly approximate the initial model. Among well-known surrogate approaches, the LIME approach \cite{ribeiro2016should} uses "kernel smoothing" as a local model approximation. The method is very sensitive to the kernel bandwidth parameter used for smoothing, which results in an unstable behavior, highly dependent on the parametrization set by the user. 

\subsubsection{Model specific methods}
The packages \textbf{xgboostExplainer}\footnote{http://github.com/AppliedDataSciencePartners/xgboostExplainer} and \textbf{randomForestExplainer}\footnote{https://github.com/ModelOriented/randomForestExplainer} are well-known tools dedicated to tree ensemble model interpretability. For a given input, the idea is to analyze the contribution of each feature to the final decision of the model by measuring in which proportion it changes the log odds of the outcome. This type of explanation is similar to the output of sensitivity analysis methods that mostly analyze the influence of each feature/groups of features on the model decision \cite{lundberg2017unified}. However, for fault/anomaly detection, the model tends to look at the same characteristics whatever the class of the input data ("normal" or "abnormal"). Instead, it would be better to characterize an anomaly in terms of abnormal values in the input characteristics. CF techniques take this approach by looking at what has changed in the input measurements/characteristics compared to a data classified as "normal" by the model. 

Over the past few years, the literature has grown extremely rich regarding CF example-based approaches \cite{singla2021explaining, goyal2019counterfactual, akula2020cocox, forster2020evaluating}. We review a certain number of them according to their categorization as model agnostic or model specific. The first category of methods consider the decision model as a black-box, and, as such, can be applied to any type of model. The second category uses the particularities of the decision model -- most often its parameters -- in order to derive an interpretation of its decisions.

\subsection{CF example-based interpretability methods}

\subsubsection{Adversarial methods}

Adversarial-based methods are used to build a virtual data which is as close as possible to the initial query data while being classified in an other class. As it is done in CF approaches, an anomaly can be explained by looking at the minimal adversarial "change" in the abnormal data that causes the classifier to "put it back" in the normal class \cite{wachter2017counterfactual, grath2018interpretable}.

Adversarial approaches require the model to be differentiable with respect to its input, and, as such, cannot be directly applied to tree ensemble models. The decision of tree ensemble models is indeed computed as a linear combination of multi-dimensional interval functions, as shown in the sec.~\ref{sec:geom_carac} of the paper (eq.~\ref{eq:dec_function}). The gradient of such an expression with respect to the input data is null inside the support of the multi-dimensional intervals, and is not defined on the interval boundaries, where the decision function is discontinuous. Approaches exist which seek to construct derivable surrogates of a tree-ensemble model: in \cite{lucic2019focus} the author replace the decision trees splitting thresholds with sigmoid functions, but this approach also suffers from vanishing gradient problem inside the supports of multi-dimensional intervals which coincide with the sigmoid tails. On the contrary, the gradient will be very strong on the interval boundaries, making it hard to adjust a learning rate which produces an adversarial example not too far from the original example.

The main problem with adversarial approaches is that they lack an explicit control of the distortion in terms of Euclidean distance between the adversarial example and the initial data point. Unlike the CF approach described in this paper, there is no guarantee to find the closest virtual example from the other class. How close is the example found will strongly depend on the learning rate and the size of the gradient step. In practice, we show in the experimental section that adversarial example-based model decision explanation methods exhibit much less stability than CF example-based ones. To evaluate the stability, we use a criterion assessing the robustness of an explanation method developed in \cite{alvarez2018robustness}, which is based on the idea that a small change in the input data should also produce a small variation in the explanation. An other assessment of the quality of the explanation is sparsity, i.e. the number of features that changed in the CF/adversarial example compared to the initial query point \cite{mothilal2020explaining}. CF methods are shown to be generally better than adversarial ones on this respect \cite{barredo2020plausible}, though there are mechanisms designed for adversarial methods, such as adding L1-regularization on the additive adversarial noise \cite{moore2019explaining}, that allow to enforce sparsity in a certain extent. The DiCE method \cite{mothilal2020explaining} also uses this approach, while also enforcing a diversity constraint on the CF/adversarial examples found.

A last advantage of the CF approach presented in the paper over adversarial ones is the possibility to find several CF examples lying inside a given radius of the query, and, also, the possibility to characterize geometrically CF decision regions that contain CF examples located within a user-defined radius. The added-value of doing so is that we can retrieve CF examples that exhibit diversity since belonging to different decision regions. It is then possible to show a panel of possible explanations to the user, though the more likely explanation is probably the simplest, i.e. the one given by the true CF example.

\subsubsection{CF model agnostic interpretability methods}
Among other model-agnostic approaches that were tested with reported results on tree ensemble models, we can mention the LORE \cite{guidotti2018local}, LEAFAGE \cite{adhikari2018example}, and CLEAR \cite{white2019measurable} approaches. The LORE approach uses local interpretable surrogates in order to derive sets of CF rules. The CLEAR method uses a technique based on feature-perturbation to discover a CF example. The LEAFAGE uses example-based reasoning to provide explanations. In particular, it learns local a linear surrogate to estimate the decision boundary of the model around a particular query point, and, then, formulates an explanation based on both CF and same-class-as-query-point examples.
Though being agnostic to the type of model, these methods rarely allow to discover close/optimal CF examples as shown experimentally in \cite{fernandez2020random} in the case of random forest models. They also suffer from the problems exhibited by surrogate methods.

\subsubsection{CF model specific interpretability methods}
In this paper, we derive a model specific CF approach, dedicated to the interpretability of tree-ensemble models decisions. As such, we narrow down the search to works that seek to exploit the specificity of decision trees / tree ensemble models to compute exact or approximate CF examples. In \cite{carreira2021counterfactual}, the authors propose a decomposition technique to obtain the decision regions of oblique trees with the purpose of performing CF explanation. This approach, though exact and totally similar in idea to the method exposed in this paper, only applies to single decision tree models, and not to tree ensemble models, limiting its potential use in complex classification/regression scenario. 
The closest work we could find to the algorithm proposed in this paper is a method to determine optimal CF sets for random forests, proposed in \cite{fernandez2020random}. This method is an improvement in terms of scalability and ability to discover optimal CF examples over two other (quasi)-exact CF approaches: MACE \cite{karimi2020model} and Feature-Tweaking (FT) \cite{tolomei2017interpretable}. The method serves the same purpose as ours, i.e. determining a set of CF examples that lie within a given radius around the query, including the best CF example, i.e. the closest in distance to the query point (hence the name "optimal CF sets" given by the authors to their approach). The algorithm presented by the authors is different from the approach presented in this paper. It is based on the idea that an equivalent decision tree can be constructed from the set of trees composing a random forest (RF) model. They evolve their method from the naive approach which consists in starting from the first tree of the RF, then, adding the second tree in all the leaves of that tree, and repeating that process until all the trees in the RF model have been added. There are unfortunately scalability issues with that approach which prevented us from testing it on our experimental scenarios, though the authors presented several optimization for pruning out unfeasable paths, in proportion as new trees are added.

Our approach decomposes the decision regions of a tree ensemble model in an exact and explicit way, by building the decision regions which are within a given range of the query dimension by dimension. As more dimensions are added, more and more regions which are outside the defined range are pruned out. This width-first approach is combined with a depth-first one traversing all the dimensions without the need to build all the regions. This strategy allows to find very quickly eligible CF regions which are used to progressively refine the search range, provoking a tremendous speed up of the branch-and-bound search process. This is this last aspect, which is by design not implementable in the approach proposed in \cite{fernandez2020random}, that ensures the scalability of our method to arbitrary large models.


\section{Problem statement}

In this article, we deal with the interpretability of tree ensemble classification models, i.e. models consisting in a collection of decision trees whose individual scores are aggregated to form the model decision.

\subsection{General problem formulation}
Given an input point and a classification model, we seek an explanation of the model prediction in that input point by comparing the latter to the nearest virtual point belonging to another class than the predicted one. We call it "virtual" since this point does not necessarily refers to an existing point in the training set (most of the time it is not), but to a point which is built "artificially" by using the model parameters (and not the training data). This approach refers to a definition of interpretability given in the sec.~6.1 of \cite{molnar2019}, and based on the idea of "counterfactual explanations": \textit{A counterfactual explanation of a prediction describes the smallest change to the feature values that changes the prediction to a predefined output.} This approach is particularly useful in fault detection scenarios, where we want to detect subtle and potentially multivariate changes in the input data that caused the data to be classified as "faulty". In such scenarios, the diagnosis provided by a CF explanation methods consists in describing what changed numerically in the input features compared to the closest data classified as "normal". 

In this paper, we provide an \textbf{exact} algorithmic solution to the problem of determining the closest CF instance to a given query for tree ensemble classification models. We also introduce a straightforward extension to tree ensemble regression models, which is presented and showcased in the experimental section. We describe several algorithmic improvements which render our approach applicable to arbitrarily large models.

The core of the approach is an algorithmic method to compute a geometrical characterization of the decision space of a tree ensemble model as a collection of multi-dimensional intervals, where each interval corresponds to a region of the input feature space inside which the model produces a uniform decision score.

\begin{figure}
\centering
\includegraphics[width=0.28\textwidth]{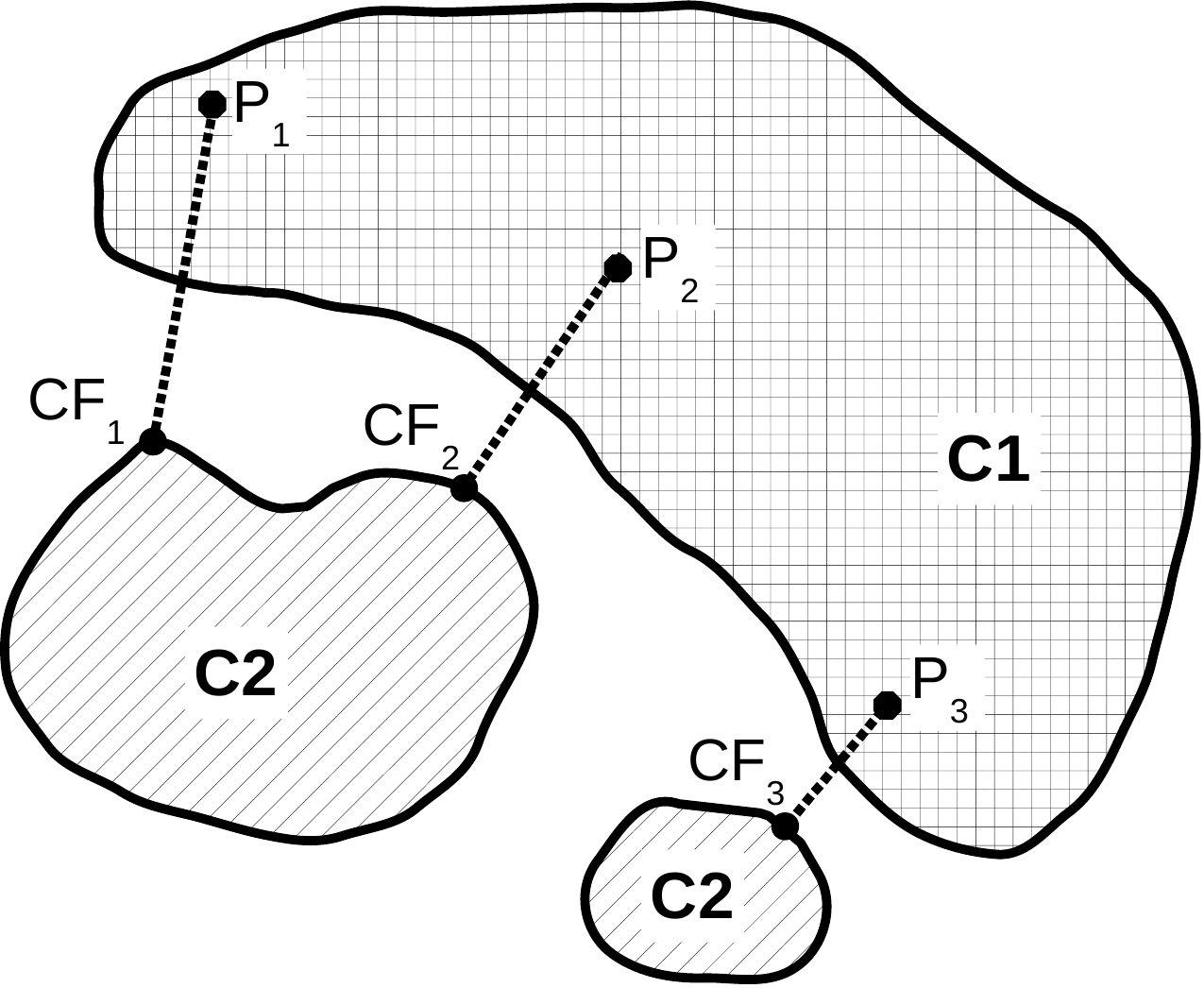}
\caption{\label{fig:CF} Illustration of the idea of CF example on a binary classification problem (C1 vs. C2). The diagram of the figure represents the decision regions of the model for the two classes C1 and C2 in the input feature space. For three points belonging to class C1, $\left(P_1, P_2, P_3\right)$, we look for the closest points belonging to the decision region of the model corresponding to class C2. These points are referred respectively as $\left(CF_1, CF_2, CF_3\right)$ and correspond to the CF examples associated respectively with $\left(P_1, P_2, P_3\right)$.}
\end{figure}

This reformulation allows us to provide an exact answer to the CF query: "given a classification model $F$ and an input point $X$ belonging to a class "$i$", i.e. such that $\textnormal{Class}\left(F(X)\right) = i$, what is the virtual point $Y$ of class "$j$" ($j \neq i$) which is the closest to $X$ in terms of Euclidean distance in the input feature space ?". The point $Y$ is called the closest CF example of class $j$ associated with $X$. In the following, we use the notation CF($X$, $j$) to refer to $Y$. The diagram of the fig.~\ref{fig:CF} illustrates the concept of closest CF example.

Stated mathematically and using the above notations, we solve the following problem:
\begin{align} \label{eq:request}
\textnormal{CF}(X, j) = \arg \min_{\left\{Y \left| \textnormal{Class}(F(Y)) = j \right.\right\}} \norm{Y - X}_2^2
\end{align}
The solution to this problem is not necessarily unique. Our approach is able to discover all the virtual points that are an optimal solution to it.

\subsection{Region-based formulation of the CF problem}
We use the following notations:
\begin{itemize}
    \item $D$: number of input features (dimension of input space)
    \item $K$: number of predefined classes in the training set of the model
    \item $F$: the $K$-class classification tree-ensemble model. In most tree-ensemble implementations, $F$ does not produce directly a classification index, but a multidimensional score in $\mathbf{R}^K$. We thus consider that $F$ is a mapping between the input feature space and $\mathbf{R}^K$, i.e., $F : \mathbf{R}^D \rightarrow \mathbf{R}^K$.
    \item Class: operator transforming the multidimensional score produced by a model into a class index ("Class" is almost always the "argmax" operator): $\textnormal{Class} : \mathbf{R}^K \rightarrow 1, \ldots, K$.
\end{itemize}

Solving the problem stated in eq.~\ref{eq:request} requires knowing the decision regions of the model $F$ in the input feature space, i.e. the regions inside which the model makes one and only one decision regarding the classification of the elements in this region. These regions are called "pure regions" of the model.

\begin{definition}[Pure region]
We say that $W \subset \mathbf{R}^D$ is a pure region of $F$ if $F$ is constant over that region, i.e.:
$$
\exists S_W \in \mathbf{R}^K \textnormal{ such that } \forall Z \in W, \; F(Z) = S_W
$$
\end{definition}
If $\textnormal{Class}\left( S_W \right) = k$, $W$ is called a pure region of class $k$ associated with the model $F$. We then write $\textnormal{Class}\left( W \right) = k$.

Characterizing the decision space of a model $F$ amounts to decomposing the domain of $F$ into distinct pure regions, which leads to the following definition:
\begin{definition} \label{def:dec_pure}
Given a model $F$ with domain $E_F \subset \mathbf{R}^D$, a decomposition of $F$ into $M$ pure regions is a decomposition $\left\{W_q \left(F \right)\right\}_{q=1, \ldots, M}$ such that:
\begin{itemize}
\item $\forall q \in \{1, \ldots, M\}, \; W_q \left(F \right)$ is a pure region of $F$
\item $\bigcup_{q=1}^{M} W_q \left(F \right) = E_F$
\item $\forall (i, j) \in \{1, \ldots, M\} \textnormal{ such that } i \neq j, \; W_i \cap W_j = \emptyset$
\end{itemize}
\end{definition}
This decomposition is not unique. For instance, each pure region in a given decomposition could be further split into several sub-regions, and the sub-regions would still form a pure decomposition in the sense of def.~\ref{def:dec_pure}.

In order to compute the minimization required by \ref{eq:request} we need to define the minimal distance between an input point $X$ and a region of the domain $E_F$ given by the model $F$.
\begin{definition} \label{def:d2region}
The minimal distance from a point $X \in \mathbf{R}^D$ to a region $\Omega \subset \mathbf{R}^D$ is defined as:
$$
dist\left( X, \Omega \right) = \min_{Y \in \Omega} \norm{Y - X}_2^2
$$
\end{definition}
$Y$ is then the point of the $\Omega$ region closest in Euclidean distance to $X$.

\begin{proposition} \label{prop:CF_pure}
Given a model $F$ and a pure region decomposition, $\left\{W_q \left(F \right)\right\}_{q=1, \ldots, M}$, the CF example of class $j$ associated with a point $X$ as defined by eq.~\ref{eq:request} is computed by solving the problem:
\begin{align}
&\textnormal{CF}(X, j) = \arg \min_{Y \in W_{q^*}} \norm{Y - X}_2^2 \textnormal{ where } \nonumber \\
&q^* = \arg \min_{q \in \left\{l \in \{1, \ldots, M \} \left| \textnormal{Class}\left( W_l \right) = j \right. \right\}} dist\left( X, W_q \left(F \right) \right) \nonumber
\end{align}
\end{proposition}

In the following, we first derive a geometrical reformulation of a tree-ensemble model as a collection of multi-dimensional intervals. Using this reformulation, we then describe a way to construct a pure region decomposition characterizing the decision space of a tree-ensemble model $F$ in an exact and exhaustive way. The pure regions are themselves constructed as multi-dimensional intervals, allowing us to compute the distance of a query point to a pure region of $F$ (using the definition of distance of def.~\ref{def:d2region}) in an exact and efficient way. Having solved the above-mentioned problems, it is then possible, according to prop.~\ref{prop:CF_pure} to determine an optimal answer to a CF query.

\section{Tree ensemble model geometrical reformulation}


\subsection{Geometrical characterisation of a decision tree} \label{sec:geom_carac}
We consider in this study binary decision trees in which each internal node represents a "test" on an input feature, i.e., is associated with an input feature index and a threshold value on this feature. This type of trees are the main component of tree-ensemble models such as XGBoost, LightGBM, ... which can be seen as a collection of such trees. In this definition, a feature index is not necessarily appearing only once inside a path between the root and a leaf of a decision tree. As in most models listed above and for the sake of generality, we allow an arbitrary number of nodes to correspond to the same feature index inside a tree path (including none). 

Given a tree of maximal depth $L$, a leaf is characterized by a succession of at most $L$ tests in each of the nodes located on the path between the root and the considered leaf. The maximal depth $L$ is an hyperparameter set by the user when specifying the model, and does not depend on the number of dimensions $D$. The same input dimension can be tested several times on a path between the root and a leaf of the tree. Still, in this configuration, there are at most two "useful" tests per input dimension, i.e. tests that do not induce any contradiction (for instance, $x > 2 \; \wedge \; x < 1$) or unnecessary checking (for instance, $x > 2 \; \wedge \; x > 1$). A tree path is indeed a succession of logical "and", thus not allowing more than a single interval test per input dimension.

Testing if an input data falls in a leaf is thus equivalent to performing interval tests "$x_d \in [a, b] $ ?" with $a \in \mathbf{R} \cup \left\{-\infty\right\}$ and $b \in \mathbf{R} \cup \left\{-\infty\right\}$ along each dimension $d$. Geometrically speaking, a decision tree leaf is therefore a multi-dimensional "box" which each face perpendicular to a certain coordinate axis. Some faces do not exist if the test interval associated to a coordinate is open on one or both sides. We thus characterize a leaf $F_i$ of a tree as a collection of $D$ intervals:
\begin{align} \label{eq:boites}
F_i = \left\{ \left[ ls_i^d, le_i^d  \right] \right\}_{d=1, \ldots, D} \textnormal{ where }
\begin{cases}
ls_i^d \leq le_i^d, \;\; \forall d \in \{1, \ldots, D \} \\
ls_i^d \in \mathbf{R} \cup \left\{-\infty\right\} \\
le_i^d \in \mathbf{R} \cup \left\{+\infty\right\}
\end{cases}
\end{align}


\subsection{Geometrical characterisation of a tree-ensemble model decision regions} \label{sec:geom_carac}
We note $f$ a decision tree and $F^f_l$ the $l$-th leaf of the tree $f$. Each leaf is associated with a pair (score, class) noted $\left(\textnormal{Score}\left[ F^f_l \right], \textnormal{Class}\left[ F^f_l \right] \right)$ with $\textnormal{Score}\left[ F^f_l \right]\in\mathbf{R}$ and $\textnormal{Class}\left[ F^f_l \right] \in 1, \ldots, K$, meaning that the leaf $F^f_l$ votes for the class $\textnormal{Class}\left[ F^f_l \right]$ with a score $\textnormal{Score}\left[ F^f_l \right]$. 

In the following, we use a vectorized notation $S^f_l \in \mathbf{R}^K$ for the score associated with a leaf $F^f_l$:
\begin{align}
S^f_l\left[ k \right] = 
\begin{cases}
\textnormal{Score}\left[ F^f_l \right] \; \textnormal{ if } k = \textnormal{Class}\left[ F^f_l \right] \\
0 \; \textnormal{ if } k \neq \textnormal{Class}\left[ F^f_l \right]
\end{cases}
\end{align}

An input data $X$ follows a unique path in a tree $f$ and thus ends in a unique leaf $F^f_l$. We then write: $f(X) = S^f_l$. Using leaf membership indicator functions, we can also write:
\begin{align} \label{eq:dec_function}
f\left(X\right) = \sum_{l=1}^L \delta_{F^f_l} \left(X \right) S^f_l \; \textnormal{ where }
\delta_{F^f_l} \left(X \right) = 
\begin{cases}
1 \textnormal{ if } X \in F^f_l \\
0 \textnormal{ if } X \notin F^f_l
\end{cases}
\end{align}

A model $F$ of type "collection of decision trees" is defined as a set of $N$ decision trees $F = \left\{ f_1, \ldots, f_N \right\}$ provided with a vector aggregation function $g : \left\{ \mathbf{R}^K \right\}^N \rightarrow \mathbf{R}^K$ used to aggregate the scores given by each decision tree composing the model into a single decision. 

Given an input feature $X \in \mathbf{R}^D$, the decision made by the model $F$ on $X$ is thus calculated by:
\begin{align} \label{eq:aggr}
F\left( X \right) = g \left( f_1 \left( X \right), \ldots, f_N \left( X \right) \right)
\end{align}
For binary classification and logistic regression models, $g$ is of the form: $\frac{1}{1 + \exp\left( - \sum_{n=1}^N f_n\left( X \right) \right)}$. For multiclass classification, $g$ is of the form $\textnormal{softmax}\left( \sum_{n=1}^N f_n\left( X \right) \right)$. For squared loss regression, $g$ is of the form $\sum_{n=1}^N f_n\left( X \right)$.

$F\left( X \right)$ is thus a vector of size $K$ containing for each class $k \in 1, \ldots, K$ the probability or the belief that $X$ belongs to the class $k$. The data $X$ is classified in the class $l \in 1, \ldots, K$ iff:
$$
\textnormal{Class}(F(X)) = \arg \max_{k \in 1, \ldots, K} F\left(X\right) \;\; = \;\; l
$$

In the following, we reformulate a tree-ensemble model $F$ as a collection of $M$ leaves/boxes $\left\{F_1, \ldots, F_M\right\}$, each associated with a vector score $S_m \in \mathbf{R}^K$. We use the notation:
\begin{align} \label{eq:spec_model}
F = \left\{ F_m, S_m \right\}_{m=1,\ldots, M} \; \textnormal{ where }
\begin{cases}
F_m = \left\{ \left[ ls_m^d, le_m^d  \right] \right\}_{d=1, \ldots, D} \\
S_m \in \mathbf{R}^K
\end{cases}
\end{align}

\section{Tree-ensemble model pure region decomposition \label{sec:decomposition}}

For a "collection of decision trees" model, we will look for a decomposition into pure regions under the form of multi-dimensional "boxes" as defined by eq.~\ref{eq:boites}. We call such a decomposition a "decomposition into pure boxes". We notice that the domain of a tree-ensemble model is exactly the union of the boxes corresponding to the leaves of the trees, leading us to build a decomposition into pure boxes of this union.

\theoremstyle{definition}
\begin{definition}
A region $W \subset \mathbf{R}^D$ is said to be a region of maximum intersection associated with a tree-ensemble model $F$ if:
\begin{itemize}
    \item there exists a subset of leaves $G = \left\{ F_{i_1}, \ldots, F_{i_L} \right\} \subset F$ such that $W \subset \bigcap_{l=1}^L F_{i_l}$.
    \item for any leaf $F_j \notin G$, $F_j \cap W = \emptyset$. 
\end{itemize}
\end{definition}

\theoremstyle{Proposition}
\begin{proposition} \label{prop:iter_maxi_pure}
A maximum intersection region associated with a tree set model is a pure region associated with that model.
\end{proposition}
Indeed, by construction, a maximal intersection region has a uniform score $S_G$ over the whole region resulting from the aggregation of the scores of the leaves that intersect to form this region, i.e., $S_G = g\left( S_{i_1}, \ldots, S_{i_L} \right)$.

Since the intersection of several boxes (multidimensional intervals) is a box as well, we call the maximum intersection region of a tree-ensemble model $F$ a box of maximum intersection associated with $F$.

\theoremstyle{definition}
\begin{definition}
Given a tree-ensemble model $F$ with $M$ leaves $\left\{F_1, \ldots, F_M\right\}$, we call a maximum intersection box decomposition of $F$ a decomposition $\left\{ B_1, \ldots, B_N\right\}$ such that:
\begin{itemize}
    \item $\forall n \in \{1, \ldots, N\}, B_n$ is a maximal intersection box associated to $F$
    \item $\bigcup_{n=1}^N B_n = \bigcup_{m=1}^M F_m$
    \item $\forall (i,j) \in \{1, \ldots, N\}$ such that $i \neq j$, $B_i \cap B_j = \emptyset$
\end{itemize}
\end{definition}

\begin{figure}
\centering
\includegraphics[width=0.45\textwidth]{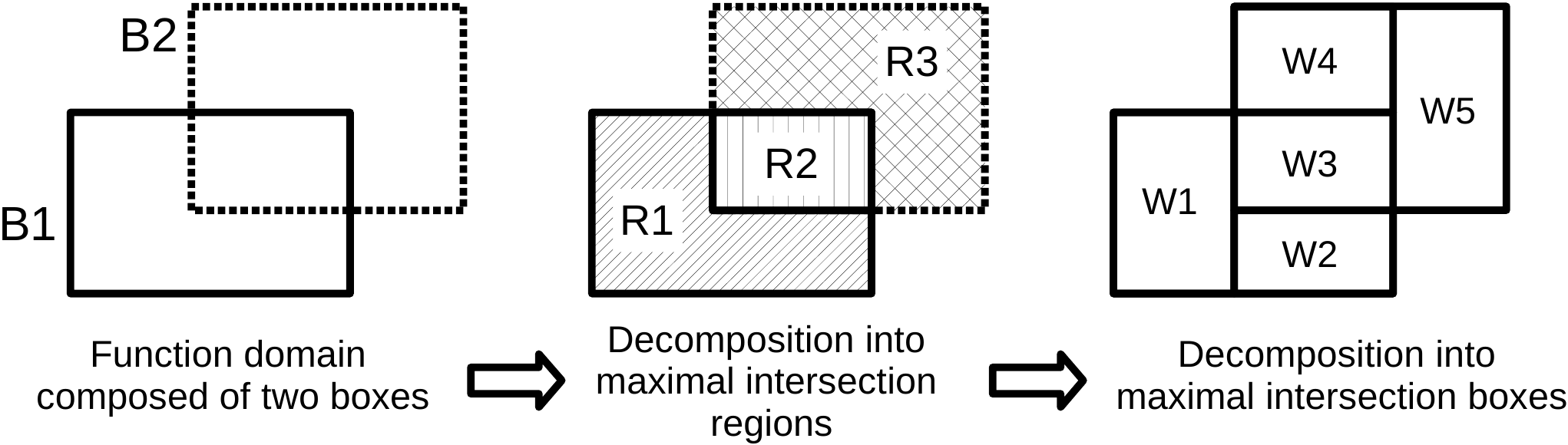}
\caption{\label{fig:boite_inter_maxi} \small{Building maximum intersection box decomposition from a union of two boxes, B1 and B2. As shown in the intermediate decomposition step into three maximum intersection regions R1, R2 and R3, the maximum intersection regions are not necessarily boxes but can be sets of of disjoint maximum intersection boxes like the region R1 which can be decomposed into two maximum intersection boxes: W1 and W2.}}
\end{figure}

As can be seen from the diagram in fig.~\ref{fig:boite_inter_maxi}, these regions are not necessarily boxes, but can also be sets of disjoint boxes, such as the regions "R1" and "R3" which are decomposed into two boxes of maximum intersection each. In general, the intersection between two $D$-dimensional boxes possesses a $2 D + 1$-boxes decomposition.

We prove the following property: 
\theoremstyle{Proposition}
\begin{proposition} \label{prop:dec_boites}
Given a tree-ensemble model $F$ with $N$ leaves with domain in $\mathbf{R}^D$, it is possible to find a pure box decomposition of the leaves of $F$ with at most $(2 N - 1)^D$ elements.
\end{proposition}
In practice, we do not need to construct the whole decision space to find the exact solution to the problem \ref{eq:request}, as exposed in the sec.~\ref{sec:branch_and_bound}, so this complexity is not a problem.

The construction is done by recurrence dimension by dimension. We prove a "stronger" result than the result of prop.~\ref{prop:dec_boites} by showing that we can construct by recurrence a decomposition into maximal intersection boxes (which are also pure regions according to prop.~\ref{prop:iter_maxi_pure}) of $F$.

\theoremstyle{Proposition}
\begin{proposition} \label{prop:rec_prop}
Given a tree-ensemble model $F$ and a maximal intersection boxes decomposition of the restriction of $F$ to the first $d$ dimensions, it is possible from the previous decomposition to build a decomposition in boxes of maximal intersection of the restriction of $F$ to the first $d+1$ dimensions (cf. def.~\ref{def:restriction_model} for the definition of "restriction" of a model)
\end{proposition}

\noindent We consider a tree-ensemble model $F$ with $N$ leaves $\left\{F_1, \ldots, F_N \right\}$ whose domain is included in $\mathbf{R}^D$. We first prove that we can construct a decomposition into boxes of maximal intersection when $D=1$ (initialization). Then we prove the recurrence property formulated in prop.~\ref{prop:rec_prop} (heredity).

\paragraph{Case $D=1$ (initialization):} The $D=1$ case is the same as finding the maximal intersections resulting from the intersection of $N$ 1D-intervals. The problem may seem combinatorial at first sight: given $m$ intervals among $N$, we have to check if these intervals have a common intersection, which, in total, makes $\sum_{m=1}^N \binom{N}{m} = 2^N - 1$ potential intersections to check.

In practice, the combinatorial aspect is broken by noticing that a maximal intersection area starts/ends in a point if and only if at least one interval among the $N$ considered starts or ends in this point. For $N$ intervals, there are at most $N$ distinct interval starts and $N$ distinct interval ends, creating at most $2N - 1$ maximum intersection areas (the last interval to terminate does not create a new maximum intersection area but terminates the last area, hence the " - 1").

An efficient algorithm to compute the maximum intersection regions consists in first sorting the $2N$ values corresponding to the beginning and ending of the $N$ intervals, and then, creating a new maximum intersection region each time an interval begins or ends. A Boolean matrix $A$ of size $N \times 2N - 1$ is created, where $A(i, j) == \textnormal{TRUE}$ means that the $i$-th interval is present in the $j$-th maximum intersection region.
This matrix can be very large, but, in general, it has a very sparse structure that allows its explicit construction whatever the number $N$ of leaves in the model.
The procedure "$Intersect1D$" described in algo.~\ref{algo:intersect_1D} implements this idea.

\SetKwInput{KwData}{Data}
\begin{algorithm}[!ht]
    \KwData{
	Collection of $N$ intervals $\left\{[s_n, e_n]\right\}_{n=1,\ldots,N}$ \\
	}
	$P \leftarrow \left\{ s_1, e_1, s_2, e_2, \ldots, s_N, e_N \right\}$ \\
	$R \leftarrow \left\{ 1, 1, 2, 2, \ldots, N, N \right\}$ \\
	$\textnormal{is\_start}[n] \leftarrow \left\{ \textnormal{\footnotesize{TRUE}}, \textnormal{\footnotesize{FALSE}}, \textnormal{\footnotesize{TRUE}}, \textnormal{\footnotesize{FALSE}}, \ldots, \textnormal{\footnotesize{TRUE}}, \textnormal{\footnotesize{FALSE}} \right\}$
	$I, V \leftarrow \textnormal{SortWithIndex}\left(P\right)$ \\
	$Q\left[n\right] \leftarrow \textnormal{is\_start}\left( R\left[ I\left[n\right] \right] \right)$ \\
	$Started[n] \leftarrow \textnormal{FALSE}$ \\
	$id.n \leftarrow 1$ \\
	$ind.unique \leftarrow \left\{ 1 \right\}$ \\
	$A[i, j] \leftarrow \textnormal{FALSE}$,  $A \in \mathcal{M}_{N \times (2N-1)}$\\
	\For{$n = 1:2N-1$}{
	    $i_n \leftarrow I[n]$ \\
	    \If{$n > 1$ and $V[n] > V[n-1]$} {
	        $id.n \leftarrow id.n + 1$ \\
	        $ind.unique[id.n] \leftarrow n$ \\
	    }
	    \If{$Q\left[i_n\right]$}{
	        $Started\left[ R\left[ i_n \right] \right] \leftarrow \textnormal{TRUE}$ \\
	    } \Else{
	        $Started\left[ R\left[ i_n \right] \right] \leftarrow \textnormal{FALSE}$ \\
	    }
	    
	    \For{$s = 1:N$}{
	        \If{$Started\left[s\right]$}{
	            $A\left[ s, id.n \right] \leftarrow \textnormal{TRUE}$ \\
	        }
	    }
	}
	
	$A \leftarrow A[:, 1:id.n]$ \\
	$V \leftarrow V[ind.unique]$ \\
	\KwResult{\\
	$\left\{ \left[bs_n, be_n\right] \right\}_{n=1, id.n}$ where $bs_n = V[n]$ and $be_n = V[n+1]$\\
	$\left\{ I_n \right\}_{n=1, id.n}$ where $I_n = \left\{ i \in \{1, \ldots, N \} \left| A\left[i, n \right] = \textnormal{TRUE} \right. \right\}$\\
	}
	
    \caption{\label{algo:intersect_1D} Procedure "$Intersect1D$" for computing the maximum intersection regions associated with $N$ 1D-intervals. The $n$-th maximal intersection region constructed is defined by the interval $\left[ bs_n, be_n \right]$ and is formed by the intersection of the intervals of the set $\left\{[s_n, e_n]\right\}_{n=1,\ldots,N}$ whose index is contained in $I_n$.}
\end{algorithm}

\paragraph{Case $D > 1$ (heredity)}
We assume that we have a decomposition in boxes of maximal intersection for a dimension $d < D$: $\left\{ B_i^d \right\}_{i=1, \ldots, L_d}$. From this decomposition, we construct a maximal intersection box decomposition at dimension $d+1$.

\noindent We introduce the notations:
\begin{itemize}
\item A box of dimension $d$ indexed by $i$ is denoted as:
$$
B_i^d = \left\{ \left[ bs_i^1, be_i^1 \right], \ldots, \left[ bs_i^d, be_i^d \right] \right\}
$$
"$bs$" stands for "box start" and "$be$" for "box end".
\item the $n$-th of the model is denoted as:
$$
F_n = \left\{ \left[ ls_n^1, le_n^1 \right], \ldots, \left[ ls_n^D, le_n^D \right] \right\}
$$
"$ls$" stand for "leaf start" and "$le$" for "leaf end".
\end{itemize}

\noindent We set the following definitions:
\theoremstyle{definition}
\begin{definition}
Given a box $B_i^D = \left\{ bs_i^l, be_i^l \right\}_{l=1, \ldots, D} \subset \mathbf{R}^D$, we call restriction of $B_i^D$ to the $d$ first dimensions ($d \leq D$) the box $B_i^d = \left\{ bs_i^l, be_i^l \right\}_{d=1, \ldots, d} \subset \mathbf{R}^d$.
By analogy, we note $F_n^d$ the restriction of the leaf $F_n$ to the first $d$ dimensions.
\end{definition}
Also:
\begin{definition} \label{def:restriction_model}
Given a tree-ensemble model $F$, we call restriction of $F$ to the first $d$ dimensions the restriction of the set of boxes corresponding to the leaves of $F$ to the first $d$ dimensions. We use the notation $Restrict_{1\rightarrow d}(F)$.
\end{definition}

The dimension-by-dimension decomposition uses the fact that we can build the intersection boxes $B_k^{d+1}$ from the maximal intersection boxes $B_i^d$ considered independently from each other.

We use the following notations:
\begin{itemize}
\item We denote $I_i^d$ the set containing the indexes of the leaves restricted to the first $d$ dimensions that intersect with the box $B_i^d$. 
\item We denote $\left\{ \left[ bs_{i,q}^{d+1}, be_{i,q}^{d+1} \right] \right\}_{q \in 1, \ldots, N_i^{d+1}}$ the decomposition into maximal intersection intervals of the subset $I_i^d$ of leaves restricted to the dimension $d+1$, i.e., of the intervals $\left\{ \left[ ls_n^{d+1}, le_n^{d+1} \right] \right\}_{n \in I_i^d}$. The notation $N_i^{d}$ refers to the cardinal of the set $I_i^d$.
\end{itemize}

The decomposition $\left\{ \left[ bs_{i,q}^{d+1}, be_{i,q}^{d+1} \right] \right\}_{q \in 1, \ldots, N_i^{d+1}}$ is obtained using the algo.~\ref{algo:intersect_1D}:
\begin{align}
&\left\{ \left[ bs_{i,q}^{d+1}, be_{i,q}^{d+1} \right] \right\}_{q \in 1, \ldots, N_i^{d+1}} \;, \; I_{i, q}^{d+1} \;\;\; \leftarrow \nonumber\\
& \;\;\;\;\;\;\;\;\;\;\;\;\;\;\; Intersect1D(\left\{ \left[ ls_n^{d+1}, le_n^{d+1} \right] \right\}_{n \in I_i^d}) \nonumber
\end{align}

In the following, we denote: $$B_{i, k}^{d+1} = \left\{B_i^d, \left[ bs_{i,k}^{d+1}, be_{i,k}^{d+1} \right] \right\}$$. 

We prove the following two properties:
\theoremstyle{Proposition}
\begin{proposition} \label{prop:is_inter_max}
The boxes $B_{i, k}^{d+1}$ are maximum intersection boxes associated with the restriction of $F$ to the first $d+1$ dimensions.
\end{proposition}

\theoremstyle{Proposition}
\begin{proposition} \label{prop:rec_relation}
The decomposition $\left\{ \left\{ B_{i, k}^{d+1} \right\}_{k \in 1, \ldots, N_i^{d+1}} \right\}_{i \in 1, \ldots, L_d}$ obtained from all maximum intersection boxes $B_i^d$ is a decomposition into maximum intersection boxes of the leaves restricted to the first $d+1$ dimensions.
\end{proposition}

The proofs of the propositions \ref{prop:is_inter_max} and \ref{prop:rec_relation} are given in the appendices.

The decomposition $\left\{ \left\{ B_{i, k}^{d+1} \right\}_{k \in 1, \ldots, N_i^{d+1}} \right\}_{i \in 1, \ldots, L_d}$ is then "flattened" into a decomposition of the form $\left\{ B_j^{d+1} \right\}_{j \in 1, \ldots, L_{d+1}}$ where $L_{d+1} = \sum_{i=1}^{L_d} N_i^{d+1}$.

A decomposition into maximum intersection boxes, and, a fortiori, a decomposition into pure boxes of the model can thus be constructed dimension by dimension by building a hierarchical tree-like structure whose nodes at depth $d$ contain the pure regions $B_i^d$. The latter form a pure region decomposition of the leaves restricted to the first $d$ dimensions. The procedure described in algo.~\ref{algo:intersect_all} implements the decomposition into maximum intersection boxes.

\SetKwInput{KwData}{Data}
\begin{algorithm}[!ht]
	\KwData{
	Tree-ensemble model $F$ specified as:
	\begin{itemize}
	    \item $N$ leaves $\left\{ B_n = \left\{ \left[ls_n^d, le_n^d \right] \right\}_{d=1,\ldots,D} \right \}_{n=1, \ldots, N}$ \\
	    \item $N$ leaf scores $S = \left\{ S_1, \ldots, S_N \right\}$ avec $\forall i, \; S_i \in \mathbf{R}^K$ \\
	    \item Aggregation function $g: \left\{ \mathbf{R}^K \times \ldots \times \mathbf{R}^K \right\} \rightarrow \mathbf{R}^K$ \\
	\end{itemize}
	}
	\KwResult{
	Set of pure boxes $R$\\
	Set of associated scores $S_R$ \\
	}
	$Tasks \leftarrow EmptyStack$ \\
	Define task $T_0 = \left\{ \left\{ \left[ls_n^1, le_n^1 \right] \right \}_{n=1, \ldots, N} \right\}$ \\
	$T_0\$I \leftarrow \left\{ 1, \ldots, N \right\}$ \\
	$T_0\$depth \leftarrow 1$ \\
	$T_0\$R \leftarrow \left\{ \right\}$ \\
	$Tasks.push(T_0)$ \\
	$R \leftarrow \emptyset$ \\
	$S_R \leftarrow \emptyset$ \\
	\While{not Tasks.isEmpty()}{
	    $T_{cur} \leftarrow Tasks.popHead()$ \\
	    $d \leftarrow T_{cur}\$depth$ \\
        $\left\{ \left[ bs_i, be_i \right] \right\}_{i \in 1, \ldots, L}, \left\{I_i \right\}_{i \in 1, \ldots, L} \leftarrow Intersect1D\left( T_{cur} \right)$ \\
        \For{$i \in 1, \ldots, L$}{
            $J_i \leftarrow T_{cur}\$I\left( I_i \right)$ \\
        	\If{$d < D$} {
                $T_i \leftarrow \left\{ \left\{ \left[ls_j^d, le_j^d \right] \right \}_{j \in J_i}  \right\}$ \\
                $T_i\$I \leftarrow J_i$ \\
                $T_i\$depth \leftarrow d+1$ \\
                $T_i\$R \leftarrow \left\{ T_{cur}\$R, \; \left[ bs_i, be_i \right] \right\}$  \\
                $Tasks.push(T_i)$ \\
            } \Else{
                $R \leftarrow \left\{ R, \; \left\{ T_{cur}\$R, \; \left[ bs_i, be_i \right] \right\} \right\}$ \\
                $S_R \leftarrow \left\{ S_R, g\left( \left\{ S\left( J_i \right) \right\} \right) \right\}$ \\
            }
        }
	}
    \caption{\label{algo:intersect_all} Procedure to decompose the leaves of a tree-ensemble model $F$ into pure boxes.}
\end{algorithm}

\begin{figure*}
\centering
\includegraphics[width=0.7\textwidth]{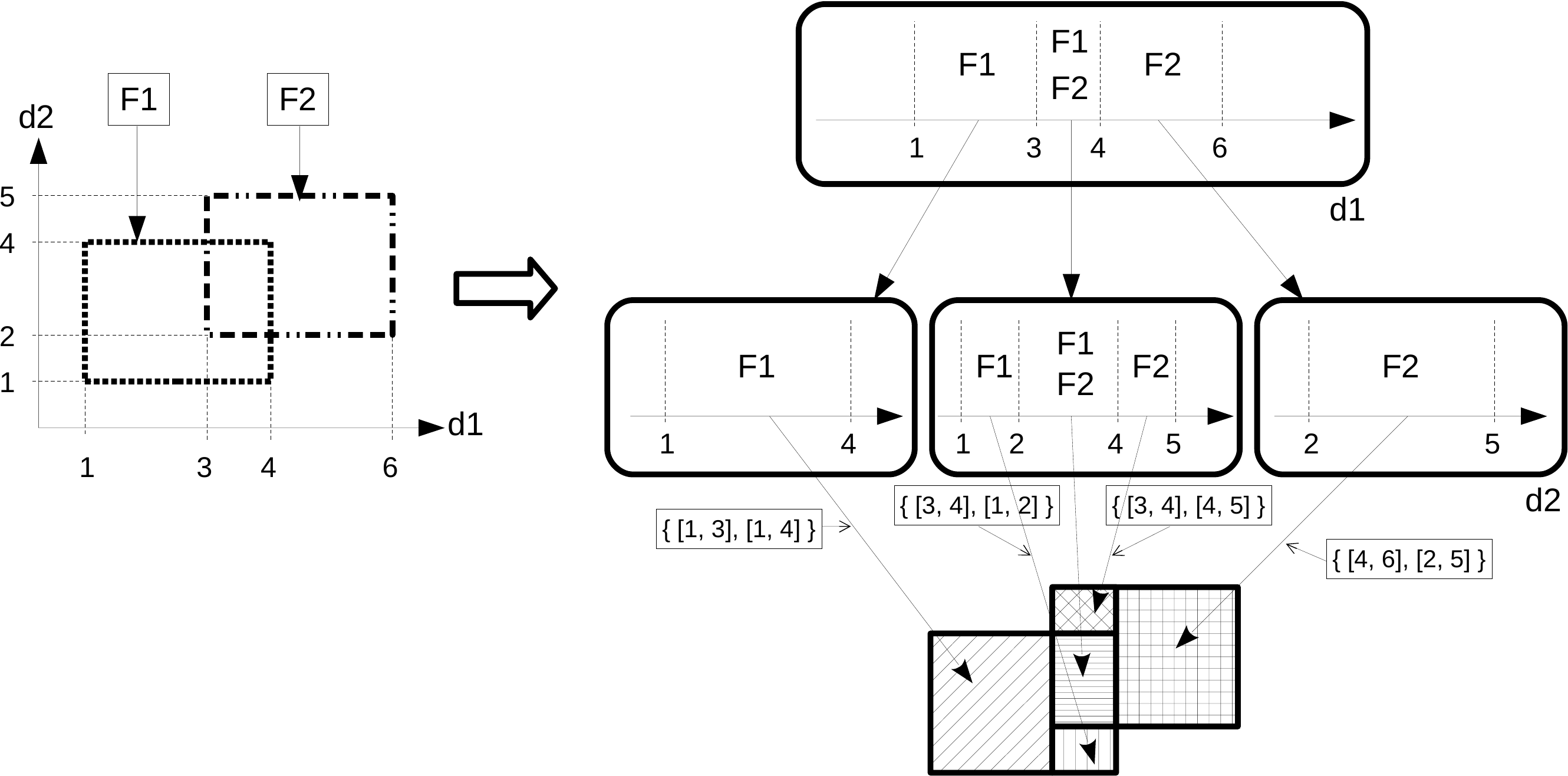}
\caption{\label{fig:intersect_all} \small{
Example of search tree built by the algo.~\ref{algo:intersect_all}. We consider here a model with two leaves, (F1, F2), in a two-dimensional feature space. The algo.~\ref{algo:intersect_all} proceeds dimension by dimension. The first stage of the search tree corresponds to the dimension d1 and to the application of the procedure \ref{algo:intersect_1D} on the leaves restricted to this dimension. The algo.~\ref{algo:intersect_1D} is then applied independently in each of the elementary intervals formed in dimension d1 to the leaves contained in each of these intervals and restricted to dimension d2. Each of the rounded-edge rectangles symbolizes an execution of the algo.~\ref{algo:intersect_1D} in the dimension associated with the considered level of the search tree and on the subset of leaves propagated from the above level and restricted to this dimension.
By following the search tree paths from the first to the last dimension, we extract a decomposition into pure box-like regions of the leaves F1 and F2. The five extracted pure boxes are represented by the hatched rectangles.
}}
\end{figure*}

The diagram in fig.~\ref{fig:intersect_all} represents graphically the tree-like structure built by the algo.~\ref{algo:intersect_all} on a simple example with two leaves in a  two-dimensional feature space. This tree-like structure is referred to as "search tree" in the remaining part of the paper. It should not be confused with the original trees composing the tree ensemble model.

\section{Answering a CF request}

\subsection{General formulation}

Given a query $X$ and a target class $j$, we seek to determine the CF example CF($X$, $j$) from the decomposition into pure boxes produced by the algo.~\ref{algo:intersect_all}. By definition, the CF example is situated on the outer envelope of a region of the target class\footnote{for any point $Y$ lying inside a region $R$, it is possible to find a point on the envelope of $R$ that is closer to the query $X$ than the point $Y$. For example, we can consider the point situated at the intersection of $[X, Y]$ with the envelope of $R$.}.

For a box $B_i$, we can easily compute the minimal distance from $X$ to $B_i$ as well as the point $Z$ of the surface of $B_i$ being at this minimal distance. The algo.~\ref{algo:dist2box} shows a way to compute the point $Z = \arg \min_{Y \in B_i} dist(X, Y)$ dimension by dimension.

\SetKwInput{KwData}{Data}
\begin{algorithm}[!ht]
	\KwData{
	$B_i = \left\{ \left[bs_i^d, be_i^d \right] \right\}_{d=1,\ldots,D}$ \\
	$X \in \mathbf{R}^D$ \\
	}
	\KwResult{
	$Z \in \mathbf{R}^D \textnormal{ such that } Z = \arg \min_{Y \in B_i} dist(X, Y)$ \\
	}
	$dist_{X \rightarrow B_i} \leftarrow 0$ \\
    \For{$d = 1:D$} {
        \If{$X[d] \in \left[ bs_i^d, be_i^d \right]$}{
            $Z[d] \leftarrow X[d]$
        } \Else{
            \If{$X[d] < bs_i^d$}{
                $Z[d] \leftarrow bs_i^d$ \\
                $dist_{X \rightarrow B_i} \leftarrow dist_{X \rightarrow B_i} + (bs_i^d - X[d])^2$ \\
            } \Else{
                $Z[d] \leftarrow be_i^d$ \\
                $dist_{X \rightarrow B_i} \leftarrow dist_{X \rightarrow B_i} + (X[d] - be_i^d)^2$ \\
            }
        }
    }
    \caption{\label{algo:dist2box} Procedure for computing $Z = dist(X, B_i)$ (def.~\ref{def:d2region}) when $B_i$ is a $D$-dimensional box. At each iteration "$d$", $dist_{X\rightarrow B_i}$ is the minimal distance between $X$ and the box $B_i$ restricted to the first $d$ dimensions, i.e., $dist_{X\rightarrow B_i} = dist\left(X, B_i^d\right)$ where $B_i^d = \left\{ \left[bs_i^l, be_i^l \right] \right\}_{l=1,\ldots,d}$.}
\end{algorithm}

Being able to compute the distance to a box dimension by dimension has a great practical interest: for each node of the search tree built by the algo.~\ref{algo:intersect_all}, we can obtain a lower bound on the distances between the query and all the pure regions below this node. Thus, if we could compute an upper bound on the distances between the query and the CF example, we could stop the construction of the search tree in all the nodes for which the calculated lower bound exceeds the upper bound (branch-and-bound principle). 
In the next section, we derive a backtracking procedure to compute such an upper bound. Then, based on this upper bound, we propose a branch-and-bound procedure that builds only a part of the search tree, ensuring the scalability of the CF example computation algorithm to arbitrarily large tree ensemble models.

\subsection{Finding the CF example relative to a subset of input features}

In some cases, we may want to answer the CF query by allowing changes to be done only on a restricted subset of input features. This corresponds to the practical scenario where it is only realistic to change some aspects in the "real life process" that underlies data generation in order to revert to normality / non-faulty state. For instance, if we take the example of credit attribution, a user that got his credit rejected may want to know what possible minimal changes he could make so that the credit is no longer rejected, knowing that there are things that cannot be changed (such as the familial situation, the number of ongoing credits ...), and things that can (such as better watching spending, saving more money, ...). Mathematically the formulation of the restricted CF query as we call it is the following:
\begin{align} \label{eq:restricted_query}
\textnormal{CF}(X, j) = \arg \min_{\left\{Y \left| \textnormal{Class}(F(Y)) = j \; \& \; \forall l \in \Omega_\textnormal{fixed} Y_l = X_l  \right.\right\}} \norm{Y - X}_2^2
\end{align}
where $\Omega_\textnormal{fixed}$ is the set of input features whose values are fixed.
The restricted CF query can be answered by pre-computing which leaves of the model are eligible given the values of the input features that are fixed, an then, by running the algo~.\ref{algo:intersect_all} on this subset of eligible leaves restricted to the dimensions that are allowed to be modified to produce the CF example.

\section{Scalability}

To answer a query, it is not necessary to build the pure region decomposition of the entire model domain as the algo.~\ref{algo:intersect_all} does. It is possible to build only a part of the domain around the query point. This approach is particularly useful for large models where the number of pure regions explodes with the number of leaves. 

\subsection{Upper bound calculation}

\begin{figure}
\centering
\includegraphics[width=0.3\textwidth]{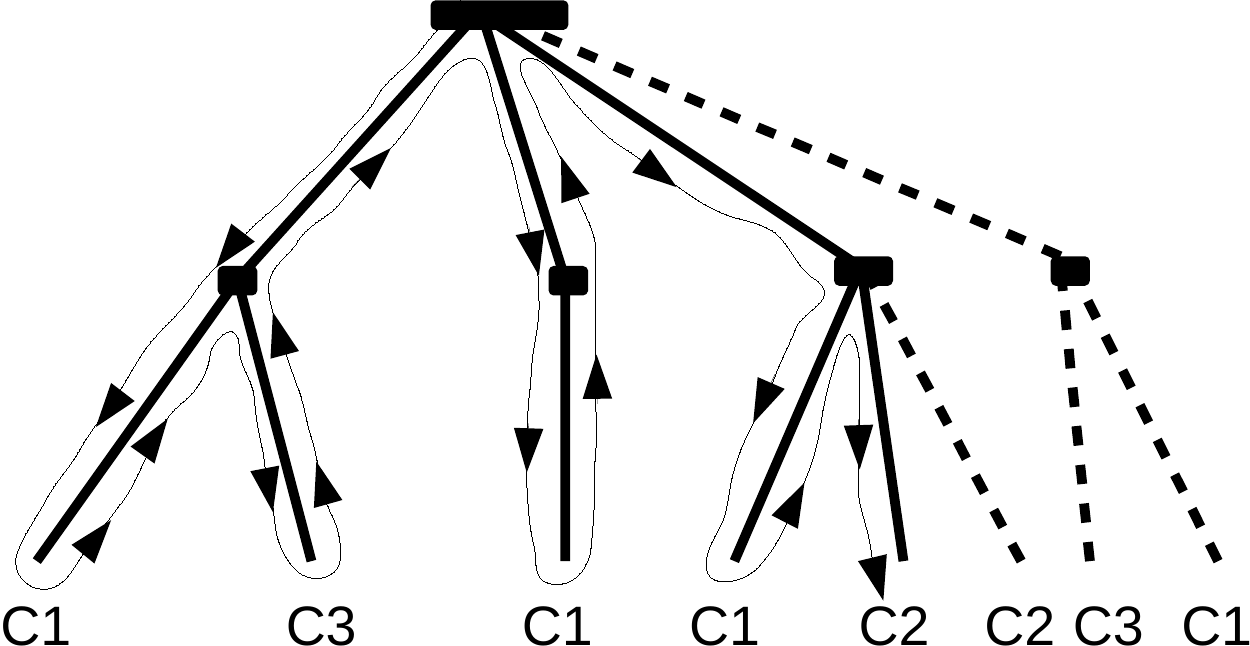}
\caption{\label{fig:backtrack} \small{High-level illustration of the backtracking procedure explained in the algo.~\ref{algo:upper_bound}. The target class is the class C2. The algorithm builds a search tree equivalent to the one built by the  algo.~\ref{algo:intersect_all} but in a depth-first manner until it finds a region belonging to C2. The build order of the search tree is indicated by the hand-drawn path. The dotted part of the structure is not constructed by the algorithm. The algo.~\ref{algo:intersect_all} in comparison performs an exhaustive construction of the search tree in a width-first manner.}}
\end{figure}

For a query point $X$ and a target class $j$, we construct only the node of the search tree at a given depth $d$ that contains the coordinate value $X[d]$. If this does not lead us to a pure region of class $j$ in the last dimension $D$, we extend the construction to the neighboring nodes by going backwards in the search tree, dimension by dimension. The corresponding backtracking approach is detailed in the algo.~\ref{algo:upper_bound}. A high level illustration of the principle of the algorithm is given in fig.~\ref{fig:backtrack}.

\SetKwInput{KwData}{Data}
\begin{algorithm}[!ht]
	\KwData{
	Tree ensemble model $F$: see spec. in algo.~\ref{algo:intersect_all} \\
	Query $X \in \mathbf{R}^D$ \\
	Target class "target" $\in 1, \ldots K$ \\ 
	}
	\KwResult{
	Upper bound $dist_{sup}$ on $dist\left(X, \textnormal{CF}(X, \textnormal{target}) \right)$ \\
	}
	
	$StackInterv[d] \leftarrow EmptyStack, \; \forall d \in 1, \ldots, D$ \\
	$Interv_0 \leftarrow \left\{ \left\{ \left[ls_n^1, le_n^1 \right] \right \}_{n=1, \ldots, N} \right\}$ \\
	$dist_0 \leftarrow 0$ \\
	$StackInterv[1].push(Interv_0, \; dist_0)$ \\
	$d_{cur} \leftarrow 1$ \\
	\While{\textnormal{TRUE}}{
	    \If{$StackInterv[d_{cur}].isEmpty()$}{
	        \If{$d_{cur}==1$} {
	            $Found \leftarrow \textnormal{FALSE}$ \\
	            \bf{return} \\
	        }
	        $d_{cur} \leftarrow d_{cur} - 1$ \\
	    } \Else{
	    
    	    $Interv_{cur}, \; dist_{cur} \leftarrow StackInterv[d_{cur}].popHead()$ \\
    	    $\left\{ \left[ bs_i, be_i \right]\right\}_{i \in 1, \ldots, L}, \left\{I_i \right\}_{i \in 1, \ldots, L} \leftarrow Intersect1D\left( Interv_{cur} \right)$ \\
    	    $\left\{ r_i \right\}_{i \in \left\{1, \ldots, L \right\}} \leftarrow \left\{ dist\left( X(d_{cur}), \left[ bs_i, be_i \right] \right) \right\}_{i \in \left\{1, \ldots, L \right\}}$ \\
    	    $i_1, i_2, \ldots, i_L \leftarrow \textnormal{sortIndex}\left( \left\{ r_i \right\}_{i \in \left\{1, \ldots, L \right\}} \right)$ \\
    	    
    	    \If{$d_{cur}==D$}{
    	        \For{$l = 1:L$} {
        	        $C_{i_l} \leftarrow \textnormal{Classe}\left( g\left( \textnormal{Score}\left( I_{i_l} \right) \right) \right)$ \\
        	        \If{$C_{i_l} == \textnormal{target}$} {
        	            $Found \leftarrow \textnormal{TRUE}$ \\
        	            $dist_{sup} \leftarrow dist_{cur} + r_{i_l}$ \\
        	            \bf{return} \\
        	        }
    	        }
    	        $StackInterv[d_{cur}].clear()$ \\
    	        $d_{cur} \leftarrow d_{cur} - 1$ \\
    	    } \Else{
        	    $d_{cur} \leftarrow d_{cur}+1$ \\
        	    \For{$l = 1:L$} {
            	    $Interv_l \leftarrow \left\{ \left\{ \left[ls_j^{d_{cur}}, le_j^{d_{cur}} \right] \right \}_{j \in I_{i_l}}  \right\}$ \\
        	        $StackInterv[d_{cur}].push(Interv_l, \; dist_{cur} + r_{i_l}^2)$ \\
        	    }
    	    }
	    
	    }
	}
	
    \caption{\label{algo:upper_bound} Backtracking procedure for computing an upper bound on the distance between the query and the closest CF example.}
\end{algorithm}

Once a target region of class $j$ is found, the distance of this region to the query point provides an upper bound on the distance of the query to the CF example of class $j$.


\subsection{Branch-and-bound formulation for the CF example search \label{sec:branch_and_bound}}

Using the upper bound computed using the algo.~\ref{algo:upper_bound}, we perform a search by constructing only the pure regions that are at a distance from the query less than or equal to the upper bound.

The procedure is explained in the algo.~\ref{algo:request}. The latter is very similar to the  algo.~\ref{algo:intersect_all}. The only difference is that we abandon the exploration in a node if the distance of the pure region defined by this node to the request exceeds the upper bound computed using the algo.~\ref{algo:upper_bound}. 

The property allowing us to do so is the fact that a node at depth $d$ in the search tree corresponds to a pure decomposition of the leaves indexed by that node and restricted to the first $d$ dimensions. 

This approach bears the generic name of branch-and-bound in the literature. Another improvement is to stop the exploration in a node if none of the boxes associated with this node vote for the target region.
As new target regions are found within the current upper bound, the latter gets updated with the distance of the query to these target regions. The search space becomes thus smaller and smaller, in proportion as new target regions are discovered.

We could still think that as we add new dimensions, the number of regions to explore grows exponentially in $(2N -1)^d$. In practice, this is not the case since less and less regions becomes eligible to contain a potential CF example. Indeed, the distances of the pure regions to the query increase when adding a new dimension (or, in the worst case, stay the same), causing less and less regions to lie within the current upper bound. 
Also, as new dimensions are added, the pure regions get associated with less and less leaves, i.e. the average number of elements in the sets $I_i^d$ decreases, causing the probability of these regions to contain leaves associated with the target class to decrease as well.
With these two effects combined, the number of regions to explore does not grow exponentially but tends to stabilize and even to decrease after exploring a certain number of dimensions.

\SetKwInput{KwData}{Data}
\begin{algorithm}[!ht]
	\KwData{
	Tree ensemble model $F$: see spec. in algo.~\ref{algo:intersect_all} \\
	Query $X \in \mathbf{R}^D$ \\
	Target class "target" $\in 1, \ldots K$ \\
	Upper bound $dist_{sup}$ on $dist\left(X, \textnormal{CF}(X, \textnormal{target}) \right)$ \\
	}
	\KwResult{
	$Y_{min} = \textnormal{CF}(X, \textnormal{target})$ \\
	}
	$Tasks \leftarrow EmptyStack$ \\
	Define task $T_0$:\\
	$T_0\$interv = \left\{ \left\{ \left[ls_n^1, le_n^1 \right] \right \}_{n=1, \ldots, N} \right\}$ \\
	$T_0\$I \leftarrow \left\{ 1, \ldots, N \right\}$ \\
	$T_0\$depth \leftarrow 1$ \\
	$T_0\$Y \leftarrow \left\{ \right\}$ \\
	$T_0\$dist \leftarrow 0$ \\
	$Tasks.push(T_0)$ \\
	$dist_{min} \leftarrow +\infty$ \\
	\While{not Tasks.isEmpty()}{
	    $T_{cur} \leftarrow Tasks.popHead()$ \\
	    $d_{cur} \leftarrow T_{cur}\$depth$ \\
	    $dist_{cur} \leftarrow Tasks\$dist$ \\
        $\left\{ \left[ bs_i, be_i \right]\right\}_{i \in 1, \ldots, L}, \left\{ I_i \right\}_{i \in 1, \ldots, L} \leftarrow Intersect1D\left( T_{cur}\$interv \right)$ \\
        \For{$i \in 1, \ldots, L$}{
            $J_i \leftarrow T_{cur}\$I\left( I_i \right)$ \\
            $dist_i \leftarrow dist_{cur} + dist\left( X[d_{cur}], \left[ bs_i, be_i \right] \right)$ \\
    	    \If{$dist_i \leq dist_{sup}$ and $\exists j \in J_i \textnormal{ tq. } \textnormal{Classe}(S_j) == \textnormal{target}$} {
    	        Define task $T_i$:\\
                $T_i\$interv \leftarrow \left\{ \left\{ \left[ls_j^{d_{cur}}, le_j^{d_{cur}} \right] \right \}_{j \in J_i}  \right\}$ \\
                $T_i\$I \leftarrow J_i$ \\
                $T_i\$depth \leftarrow d_{cur}+1$ \\
                $T_i\$dist \leftarrow dist_i$ \\
                \If{$X[d_{cur}] \in \left[ bs_i, be_i \right]$}{
                    $pt_i \leftarrow X[d_{cur}]$
                } \Else{
                    \If{$X[d_{cur}] < bs_i$}{
                        $pt_i \leftarrow bs_i$
                    } \Else{
                        $pt_i \leftarrow be_i$
                    }
                }
                $T_i\$Y \leftarrow \left[ T_{cur}\$Y, \; pt_i \right]$  \\
                \If{$d_{cur} < D$} {
                    $Tasks.push(T_i)$ \\
                } \Else{
                    \If{$\textnormal{Classe}\left( g\left( \left\{ S_l \right\}_{l \in J_i} \right) \right) == \textnormal{target}$}{
                        $dist_{sup} \leftarrow dist_i$ \\
                        \If{$dist_{sup} < dist_{min}$}{
                            $dist_{min} \leftarrow dist_{sup}$ \\
                            $Y_{min} \leftarrow T_i\$Y$
                        }
                    }
                }
            }
        }
	}
	
    \caption{\label{algo:request} Branch-and-bound procedure for computing CF($X$, target). This procedure constructs only a partial decomposition of the domain of the model lying inside a hyper-sphere centered in $X$ and of radius $dist_{sup}$.}
\end{algorithm}

\subsection{Parallel formulation of the algo.~\ref{algo:intersect_all}}

The search tree can be built concurrently: the nodes at the same depth can indeed be treated independently from each other, leading to a natural parallelization of the algo.~\ref{algo:request}. The only constraint that has to be imposed when treating a given node's task is that the parent node's task has been treated. From here, implementing the concurrent version of the algo.~\ref{algo:request} is pretty straightforward: we replace the stack structure containing the tasks in each node with a concurrent stack allowing for a concurrent execution of the tasks. Thus, instead of being unstacked one after the other by the main loop of the algorithm (line 10), tasks will be unstacked and treated concurrently.

\subsection{Pre-sorting of dimensions}

We notice that the algo.~\ref{algo:intersect_all} boils down to applying the algo.~\ref{algo:intersect_1D} in each node of the search tree represented in the  fig.~\ref{fig:intersect_all} on a selection of leaves restricted to the dimension following the dimension associated with the node. Improving the performance of this elementary algorithmic brick is thus key to overall performance improvement. 
A considerable speed increase can be achieved by pre-sorting the multidimensional intervals corresponding to the model leaves in each dimension prior to applying the algo.~\ref{algo:intersect_all}. 
The latter operates on 1D-intervals corresponding to a selection of leaves restricted to one dimension, and sorts the corresponding interval ends. Since selecting a subset of elements from a sorted set does not break the sort order, the pre-sorting removes the need for the the "SortWithIndex" operation in algo.~\ref{algo:intersect_1D} - line 3. 
This simple "pre-processing" leads to a major speedup: the complexity of the algo.~\ref{algo:intersect_1D} drops from O(N.log(N)) to O(N) where "N" is the number of intervals considered in a node of the search tree.

\subsection{Using efficient data structures}

To represent the selection of leaves $I_i$ associated with a node of the search tree, we use dynamically allocated bitsets of size equal to the total number of leaves of the model. These structures are very efficient when used in the procedure \ref{algo:intersect_1D} instead of the sparse matrix $A$: adding/removing an interval (corresponding to the restriction of a leaf to the current dimension) boils down to setting/unsetting a bit.

\subsection{Using mixed width-first / depth first exploration strategies}

Using the upper bound computed with the algo.~\ref{algo:upper_bound}, we can eliminate all the leaves which are situated farther from the query point than the upper bound. 
But this is only done once at the beginning before running the exhaustive width-first strategy implemented by the algo.~\ref{algo:request}.
Instead, we could run the algo.~\ref{algo:request} till a certain depth, and, then, change to a depth first strategy using the algo.~\ref{algo:upper_bound}. The latter is perfectly valid to explore the search tree as well. The problem with depth-first approaches is that, unlike width-first approaches, they cannot be parallelized. But, unlike width-first approaches which stack problems faster than they are treated, they do not suffer from memory problems. 
The idea is to construct the search tree up to a certain depth, and, then, to define chunks of problems that serve as a start for a depth-first exploration in the remaining dimensions, allowing that way several depth-first search processes to be run in parallel. When one of the depth-first processes discovers a new target region, the upper bound on the distance to the closest CF example is updated and made available to all the other depth-first processes, allowing a quick pruning of the search tree.

\subsection{CF sets}

It is possible to store all the multi-dimensional intervals that contain a CF example situated inside a predefined upper bound. The use of this functionality is demonstrated in the use case \ref{sec:use_case_3_3}. The interest of CF sets over CF examples for model decision explanation has been discussed in \cite{fernandez2020random}. It consists in providing the user with intervals in which the input characteristics should lie in order to achieve a certain goal (in the presented use case, the goal is a given buying price for a house).

\section{Experiments} \label{sec:experiments}

Since we formally prove that the CF examples found are the closest in terms of distance to the query, we do not need to assess experimentally the optimality and validity of the CF examples found, as other CF-based methods do \cite{mothilal2020explaining} in their quantitative evaluation. We rather place ourselves on a subjective level, and try to assess the plausibility \cite{keane2021if} of the provided explanations by formulating them explicitly from the numerical representation provided by the CF algorithm (i.e., from the vector of numerical changes to apply to the original query so that it be classified in the target class). To better assess the plausibility of the explanations, we make use of visual data, trusting the human eye to decide whether the changes that were brought to the query image by the CF approach make sense or not. 

As advised in \cite{keane2021if} and when possible (non-categorical data), we assess criteria such as the stability of the obtained explanations when introducing slight changes in the query. We also evaluate the sparsity of the provided explanations, in the idea that the fewer input features get modified, the more intelligible is the explanation for a human \cite{fernandez2019relevance}. For these criteria, the comparison is performed with an adversarial method.

The first use case is a dataset of credit approval/denial trying to relate a set of exogenous features to the decision of granting/denying a credit to a customer. We first formulate the CF diagnosis in the case the credit was denied, and analyse the minimal changes the customer as to make so that the credit be granted tom him. We then simulate the case when the customer can change only a few variables, the other ones not being in his control.

The second use case applies the CF approach to miss-classified data of the MNIST digits datasets. We represent visually the minimal changes one has to make to the original image so that it be classified in the correct class. We analyse if this changes make sense visually. This dataset allows also the evaluation of stability and sparsity. In particular, it does not contain categorical variables, allowing an easy deployment of adversarial methods for comparison.

The last use case is a regression problem whose goal is to predict houses selling price from a set of exogenous variables. We use the CF approach to figure out what changes should be made to the input variables (at least the ones that can be changed) to make the selling price closer to a target price.

\subsection{Use case 1}

In this use case, we train a decision making model on a credit allocation dataset. We use a CF example-based approach to explain each of the credit denial decisions by mentioning the system's recommendation, i.e. the characteristics that should be changed a minima so that the same credit be granted to the customer.

The model predicting if a credit was granted or not is an XGBoost binary classification model with 250 trees of depth 8. The dataset contains 20 numerical and categorical characteristics such as the number of ongoing loans, the repayment history of previous loans, the amount asked, the purpose of the loan (type of purchase), the borrower's salary and savings, the type of employment contract, the number of years of seniority in the current job ...

In tab.~\ref{tab:credit_denial}, we show five examples of credit denial diagnosed using our CF approach. The tables represent the query point (client current status) and the corresponding CF example, i.e. the changes that should be applied a minima to the current status so that the credit be granted. 

\begin{table*}[ht]
\caption{\label{tab:credit_denial} \small{Decision explanations for five cases of credit denial.}}
\centering
\resizebox{0.9\textwidth}{!}{\begin{tabular}{@{}ccllccccllcl@{}}
\toprule
\rowcolor[HTML]{FFFFFF} 
\multicolumn{1}{|c|}{\cellcolor[HTML]{FFFFFF}{\color[HTML]{000000} \textbf{}}} &
  \multicolumn{1}{c|}{\cellcolor[HTML]{FFFFFF}{\color[HTML]{000000} \textbf{\begin{tabular}[c]{@{}c@{}}Amount \\ refused\end{tabular}}}} &
  \multicolumn{1}{c|}{\cellcolor[HTML]{FFFFFF}{\color[HTML]{000000} \textbf{Purpose}}} &
  \multicolumn{1}{c|}{\cellcolor[HTML]{FFFFFF}{\color[HTML]{000000} \textbf{\begin{tabular}[c]{@{}c@{}}Credit\\ history\end{tabular}}}} &
  \multicolumn{1}{c|}{\cellcolor[HTML]{FFFFFF}{\color[HTML]{000000} \textbf{\begin{tabular}[c]{@{}c@{}}Installment rate \\ in \% of \\ disposable\\  income\end{tabular}}}} &
  \multicolumn{1}{c|}{\cellcolor[HTML]{FFFFFF}{\color[HTML]{000000} \textbf{\begin{tabular}[c]{@{}c@{}}Status of \\ existing \\ checking \\ account\end{tabular}}}} &
  \multicolumn{1}{c|}{\cellcolor[HTML]{FFFFFF}{\color[HTML]{000000} \textbf{\begin{tabular}[c]{@{}c@{}}Number of \\ existing \\ credits at \\ this bank\end{tabular}}}} &
  \multicolumn{1}{c|}{\cellcolor[HTML]{FFFFFF}{\color[HTML]{000000} \textbf{\begin{tabular}[c]{@{}c@{}}Savings account\\ /bonds\end{tabular}}}} &
  \multicolumn{1}{c|}{\cellcolor[HTML]{FFFFFF}{\color[HTML]{000000} \textbf{\begin{tabular}[c]{@{}c@{}}Present \\ employment\\  since\end{tabular}}}} &
  \multicolumn{1}{c|}{\cellcolor[HTML]{FFFFFF}{\color[HTML]{000000} \textbf{\begin{tabular}[c]{@{}c@{}}Credit duration \\ in months\end{tabular}}}} &
  \multicolumn{1}{c|}{\cellcolor[HTML]{FFFFFF}{\color[HTML]{000000} \textbf{\begin{tabular}[c]{@{}c@{}}Other debtors /\\  guarantors\end{tabular}}}} &
  \multicolumn{1}{c|}{\cellcolor[HTML]{FFFFFF}{\color[HTML]{000000} \textbf{Telephone}}} \\ \midrule
\rowcolor[HTML]{FFFFFF} 
\multicolumn{1}{|c|}{\cellcolor[HTML]{FFFFFF}{\color[HTML]{000000} \textbf{Client status}}} &
  \multicolumn{1}{c|}{\cellcolor[HTML]{FFFFFF}{\color[HTML]{000000} }} &
  \multicolumn{1}{l|}{\cellcolor[HTML]{FFFFFF}{\color[HTML]{000000} }} &
  \multicolumn{1}{c|}{\cellcolor[HTML]{FFFFFF}{\color[HTML]{000000} \begin{tabular}[c]{@{}c@{}}existing credits \\ paid back duly \\ till now\end{tabular}}} &
  \multicolumn{1}{c|}{\cellcolor[HTML]{FFFFFF}{\color[HTML]{000000} 3}} &
  \multicolumn{1}{c|}{\cellcolor[HTML]{FFFFFF}{\color[HTML]{000000} $<$ 0}} &
  \multicolumn{1}{c|}{\cellcolor[HTML]{FFFFFF}{\color[HTML]{000000} 2}} &
  \multicolumn{1}{c|}{\cellcolor[HTML]{FFFFFF}{\color[HTML]{000000} $<$  100}} &
  \multicolumn{1}{l|}{\cellcolor[HTML]{FFFFFF}{\color[HTML]{000000} }} &
  \multicolumn{1}{l|}{\cellcolor[HTML]{FFFFFF}{\color[HTML]{000000} }} &
  \multicolumn{1}{c|}{\cellcolor[HTML]{FFFFFF}{\color[HTML]{000000} none}} &
  \multicolumn{1}{l|}{\cellcolor[HTML]{FFFFFF}{\color[HTML]{000000} }} \\ \cmidrule(r){1-1} \cmidrule(l){3-12} 
\rowcolor[HTML]{FFFFFF} 
\multicolumn{1}{|c|}{\cellcolor[HTML]{FFFFFF}{\color[HTML]{000000} \textbf{Recommendation}}} &
  \multicolumn{1}{c|}{\multirow{-4.8}{*}{\cellcolor[HTML]{FFFFFF}{\color[HTML]{000000} \textbf{\$ 701}}}} &
  \multicolumn{1}{l|}{\cellcolor[HTML]{FFFFFF}{\color[HTML]{000000} }} &
  \multicolumn{1}{c|}{\cellcolor[HTML]{FFFFFF}{\color[HTML]{000000} \begin{tabular}[c]{@{}c@{}}all credits at \\ this bank paid \\ back duly\end{tabular}}} &
  \multicolumn{1}{c|}{\cellcolor[HTML]{FFFFFF}{\color[HTML]{000000} at most 2.5}} &
  \multicolumn{1}{c|}{\cellcolor[HTML]{FFFFFF}{\color[HTML]{000000} 0 $\leq$ ... $<$ 200}} &
  \multicolumn{1}{c|}{\cellcolor[HTML]{FFFFFF}{\color[HTML]{000000} at most 1}} &
  \multicolumn{1}{c|}{\cellcolor[HTML]{FFFFFF}{\color[HTML]{000000} 500 $\leq$ ... $<$  1000}} &
  \multicolumn{1}{l|}{\cellcolor[HTML]{FFFFFF}{\color[HTML]{000000} }} &
  \multicolumn{1}{l|}{\cellcolor[HTML]{FFFFFF}{\color[HTML]{000000} }} &
  \multicolumn{1}{c|}{\cellcolor[HTML]{FFFFFF}{\color[HTML]{000000} co-applicant}} &
  \multicolumn{1}{l|}{\cellcolor[HTML]{FFFFFF}{\color[HTML]{000000} }} \\ \midrule
\rowcolor[HTML]{C0C0C0} 
{\color[HTML]{000000} \textbf{}} &
  {\color[HTML]{000000} } &
  {\color[HTML]{000000} } &
  {\color[HTML]{000000} } &
  \multicolumn{1}{l}{\cellcolor[HTML]{C0C0C0}{\color[HTML]{000000} }} &
  \multicolumn{1}{l}{\cellcolor[HTML]{C0C0C0}{\color[HTML]{000000} }} &
  \multicolumn{1}{l}{\cellcolor[HTML]{C0C0C0}{\color[HTML]{000000} }} &
  \multicolumn{1}{l}{\cellcolor[HTML]{C0C0C0}{\color[HTML]{000000} }} &
  {\color[HTML]{000000} } &
  {\color[HTML]{000000} } &
  \multicolumn{1}{l}{\cellcolor[HTML]{C0C0C0}{\color[HTML]{000000} }} &
  {\color[HTML]{000000} } \\ \midrule
\rowcolor[HTML]{FFFFFF} 
\multicolumn{1}{|c|}{\cellcolor[HTML]{FFFFFF}{\color[HTML]{000000} \textbf{Client status}}} &
  \multicolumn{1}{c|}{\cellcolor[HTML]{FFFFFF}{\color[HTML]{000000} }} &
  \multicolumn{1}{l|}{\cellcolor[HTML]{FFFFFF}{\color[HTML]{000000} }} &
  \multicolumn{1}{l|}{\cellcolor[HTML]{FFFFFF}{\color[HTML]{000000} }} &
  \multicolumn{1}{c|}{\cellcolor[HTML]{FFFFFF}{\color[HTML]{000000} }} &
  \multicolumn{1}{c|}{\cellcolor[HTML]{FFFFFF}{\color[HTML]{000000} $<$ 0}} &
  \multicolumn{1}{c|}{\cellcolor[HTML]{FFFFFF}{\color[HTML]{000000} }} &
  \multicolumn{1}{c|}{\cellcolor[HTML]{FFFFFF}{\color[HTML]{000000} }} &
  \multicolumn{1}{c|}{\cellcolor[HTML]{FFFFFF}{\color[HTML]{000000} 1 $\leq$ ... $<$ 4 years}} &
  \multicolumn{1}{c|}{\cellcolor[HTML]{FFFFFF}{\color[HTML]{000000} Client asked for 16}} &
  \multicolumn{1}{c|}{\cellcolor[HTML]{FFFFFF}{\color[HTML]{000000} none}} &
  \multicolumn{1}{l|}{\cellcolor[HTML]{FFFFFF}{\color[HTML]{000000} }} \\ \cmidrule(r){1-1} \cmidrule(l){3-12} 
\rowcolor[HTML]{FFFFFF} 
\multicolumn{1}{|c|}{\cellcolor[HTML]{FFFFFF}{\color[HTML]{000000} \textbf{Recommendation}}} &
  \multicolumn{1}{c|}{\multirow{-2}{*}{\cellcolor[HTML]{FFFFFF}{\color[HTML]{000000} \textbf{\$ 330}}}} &
  \multicolumn{1}{l|}{\cellcolor[HTML]{FFFFFF}{\color[HTML]{000000} }} &
  \multicolumn{1}{l|}{\cellcolor[HTML]{FFFFFF}{\color[HTML]{000000} }} &
  \multicolumn{1}{c|}{\cellcolor[HTML]{FFFFFF}{\color[HTML]{000000} }} &
  \multicolumn{1}{c|}{\cellcolor[HTML]{FFFFFF}{\color[HTML]{000000} 0 $\leq$ ... $<$ 200}} &
  \multicolumn{1}{c|}{\cellcolor[HTML]{FFFFFF}{\color[HTML]{000000} }} &
  \multicolumn{1}{c|}{\cellcolor[HTML]{FFFFFF}{\color[HTML]{000000} }} &
  \multicolumn{1}{c|}{\cellcolor[HTML]{FFFFFF}{\color[HTML]{000000} 4 $\leq$ ... $<$ 7 years}} &
  \multicolumn{1}{c|}{\cellcolor[HTML]{FFFFFF}{\color[HTML]{000000} at most 15}} &
  \multicolumn{1}{c|}{\cellcolor[HTML]{FFFFFF}{\color[HTML]{000000} co-applicant}} &
  \multicolumn{1}{c|}{\cellcolor[HTML]{FFFFFF}{\color[HTML]{000000} }} \\ \midrule
\rowcolor[HTML]{C0C0C0} 
{\color[HTML]{000000} \textbf{}} &
  {\color[HTML]{000000} } &
  {\color[HTML]{000000} } &
  {\color[HTML]{000000} } &
  \multicolumn{1}{l}{\cellcolor[HTML]{C0C0C0}{\color[HTML]{000000} }} &
  \multicolumn{1}{l}{\cellcolor[HTML]{C0C0C0}{\color[HTML]{000000} }} &
  \multicolumn{1}{l}{\cellcolor[HTML]{C0C0C0}{\color[HTML]{000000} }} &
  \multicolumn{1}{l}{\cellcolor[HTML]{C0C0C0}{\color[HTML]{000000} }} &
  {\color[HTML]{000000} } &
  {\color[HTML]{000000} } &
  \multicolumn{1}{l}{\cellcolor[HTML]{C0C0C0}{\color[HTML]{000000} }} &
  {\color[HTML]{000000} } \\ \midrule
\rowcolor[HTML]{FFFFFF} 
\multicolumn{1}{|c|}{\cellcolor[HTML]{FFFFFF}{\color[HTML]{000000} \textbf{Client status}}} &
  \multicolumn{1}{c|}{\cellcolor[HTML]{FFFFFF}{\color[HTML]{000000} }} &
  \multicolumn{1}{l|}{\cellcolor[HTML]{FFFFFF}{\color[HTML]{000000} }} &
  \multicolumn{1}{l|}{\cellcolor[HTML]{FFFFFF}{\color[HTML]{000000} }} &
  \multicolumn{1}{c|}{\cellcolor[HTML]{FFFFFF}{\color[HTML]{000000} }} &
  \multicolumn{1}{c|}{\cellcolor[HTML]{FFFFFF}{\color[HTML]{000000} $<$ 0}} &
  \multicolumn{1}{c|}{\cellcolor[HTML]{FFFFFF}{\color[HTML]{000000} 2}} &
  \multicolumn{1}{c|}{\cellcolor[HTML]{FFFFFF}{\color[HTML]{000000} $<$  100}} &
  \multicolumn{1}{l|}{\cellcolor[HTML]{FFFFFF}{\color[HTML]{000000} }} &
  \multicolumn{1}{l|}{\cellcolor[HTML]{FFFFFF}{\color[HTML]{000000} }} &
  \multicolumn{1}{l|}{\cellcolor[HTML]{FFFFFF}{\color[HTML]{000000} }} &
  \multicolumn{1}{l|}{\cellcolor[HTML]{FFFFFF}{\color[HTML]{000000} }} \\ \cmidrule(r){1-1} \cmidrule(l){3-12} 
\rowcolor[HTML]{FFFFFF} 
\multicolumn{1}{|c|}{\cellcolor[HTML]{FFFFFF}{\color[HTML]{000000} \textbf{Recommendation}}} &
  \multicolumn{1}{c|}{\multirow{-2}{*}{\cellcolor[HTML]{FFFFFF}{\color[HTML]{000000} \textbf{\$ 716}}}} &
  \multicolumn{1}{l|}{\cellcolor[HTML]{FFFFFF}{\color[HTML]{000000} \textbf{}}} &
  \multicolumn{1}{l|}{\cellcolor[HTML]{FFFFFF}{\color[HTML]{000000} }} &
  \multicolumn{1}{c|}{\cellcolor[HTML]{FFFFFF}{\color[HTML]{000000} }} &
  \multicolumn{1}{c|}{\cellcolor[HTML]{FFFFFF}{\color[HTML]{000000} 0 $\leq$ ... $<$ 200}} &
  \multicolumn{1}{c|}{\cellcolor[HTML]{FFFFFF}{\color[HTML]{000000} at most 1}} &
  \multicolumn{1}{c|}{\cellcolor[HTML]{FFFFFF}{\color[HTML]{000000} 100 $\leq$ ... $<$  500}} &
  \multicolumn{1}{l|}{\cellcolor[HTML]{FFFFFF}{\color[HTML]{000000} }} &
  \multicolumn{1}{l|}{\cellcolor[HTML]{FFFFFF}{\color[HTML]{000000} }} &
  \multicolumn{1}{l|}{\cellcolor[HTML]{FFFFFF}{\color[HTML]{000000} }} &
  \multicolumn{1}{l|}{\cellcolor[HTML]{FFFFFF}{\color[HTML]{000000} }} \\ \midrule
\rowcolor[HTML]{C0C0C0} 
{\color[HTML]{000000} \textbf{}} &
  {\color[HTML]{000000} } &
  {\color[HTML]{000000} } &
  {\color[HTML]{000000} } &
  \multicolumn{1}{l}{\cellcolor[HTML]{C0C0C0}{\color[HTML]{000000} }} &
  \multicolumn{1}{l}{\cellcolor[HTML]{C0C0C0}{\color[HTML]{000000} }} &
  \multicolumn{1}{l}{\cellcolor[HTML]{C0C0C0}{\color[HTML]{000000} }} &
  \multicolumn{1}{l}{\cellcolor[HTML]{C0C0C0}{\color[HTML]{000000} }} &
  {\color[HTML]{000000} } &
  {\color[HTML]{000000} } &
  \multicolumn{1}{l}{\cellcolor[HTML]{C0C0C0}{\color[HTML]{000000} }} &
  {\color[HTML]{000000} } \\ \midrule
\rowcolor[HTML]{FFFFFF} 
\multicolumn{1}{|c|}{\cellcolor[HTML]{FFFFFF}{\color[HTML]{000000} \textbf{Client status}}} &
  \multicolumn{1}{c|}{\cellcolor[HTML]{FFFFFF}{\color[HTML]{000000} }} &
  \multicolumn{1}{l|}{\cellcolor[HTML]{FFFFFF}{\color[HTML]{000000} }} &
  \multicolumn{1}{l|}{\cellcolor[HTML]{FFFFFF}{\color[HTML]{000000} }} &
  \multicolumn{1}{c|}{\cellcolor[HTML]{FFFFFF}{\color[HTML]{000000} }} &
  \multicolumn{1}{c|}{\cellcolor[HTML]{FFFFFF}{\color[HTML]{000000} $<$ 0}} &
  \multicolumn{1}{c|}{\cellcolor[HTML]{FFFFFF}{\color[HTML]{000000} }} &
  \multicolumn{1}{c|}{\cellcolor[HTML]{FFFFFF}{\color[HTML]{000000} $<$  100}} &
  \multicolumn{1}{c|}{\cellcolor[HTML]{FFFFFF}{\color[HTML]{000000} 1 $\leq$ ... $<$ 4 years}} &
  \multicolumn{1}{c|}{\cellcolor[HTML]{FFFFFF}{\color[HTML]{000000} Client asked for 14}} &
  \multicolumn{1}{c|}{\cellcolor[HTML]{FFFFFF}{\color[HTML]{000000} none}} &
  \multicolumn{1}{c|}{\cellcolor[HTML]{FFFFFF}{\color[HTML]{000000} none registered}} \\ \cmidrule(r){1-1} \cmidrule(l){3-12} 
\rowcolor[HTML]{FFFFFF} 
\multicolumn{1}{|c|}{\cellcolor[HTML]{FFFFFF}{\color[HTML]{000000} \textbf{Recommendation}}} &
  \multicolumn{1}{c|}{\multirow{-2}{*}{\cellcolor[HTML]{FFFFFF}{\color[HTML]{000000} \textbf{\$ 342}}}} &
  \multicolumn{1}{c|}{\cellcolor[HTML]{FFFFFF}{\color[HTML]{000000} \textbf{}}} &
  \multicolumn{1}{l|}{\cellcolor[HTML]{FFFFFF}{\color[HTML]{000000} \textbf{}}} &
  \multicolumn{1}{l|}{\cellcolor[HTML]{FFFFFF}{\color[HTML]{000000} }} &
  \multicolumn{1}{c|}{\cellcolor[HTML]{FFFFFF}{\color[HTML]{000000} $\leq$ ... $<$ 200}} &
  \multicolumn{1}{c|}{\cellcolor[HTML]{FFFFFF}{\color[HTML]{000000} }} &
  \multicolumn{1}{c|}{\cellcolor[HTML]{FFFFFF}{\color[HTML]{000000} 100 $\leq$ ... $<$  500}} &
  \multicolumn{1}{c|}{\cellcolor[HTML]{FFFFFF}{\color[HTML]{000000} 4 $\leq$ ... $<$ 7 years}} &
  \multicolumn{1}{c|}{\cellcolor[HTML]{FFFFFF}{\color[HTML]{000000} at most 13}} &
  \multicolumn{1}{c|}{\cellcolor[HTML]{FFFFFF}{\color[HTML]{000000} co-applicant}} &
  \multicolumn{1}{c|}{\cellcolor[HTML]{FFFFFF}{\color[HTML]{000000} \begin{tabular}[c]{@{}c@{}}yes, registered under \\ the customer's name\end{tabular}}} \\ \midrule
\rowcolor[HTML]{C0C0C0} 
{\color[HTML]{000000} \textbf{}} &
  {\color[HTML]{000000} } &
  {\color[HTML]{000000} } &
  {\color[HTML]{000000} } &
  \multicolumn{1}{l}{\cellcolor[HTML]{C0C0C0}{\color[HTML]{000000} }} &
  \multicolumn{1}{l}{\cellcolor[HTML]{C0C0C0}{\color[HTML]{000000} }} &
  \multicolumn{1}{l}{\cellcolor[HTML]{C0C0C0}{\color[HTML]{000000} }} &
  \multicolumn{1}{l}{\cellcolor[HTML]{C0C0C0}{\color[HTML]{000000} }} &
  {\color[HTML]{000000} } &
  {\color[HTML]{000000} } &
  \multicolumn{1}{l}{\cellcolor[HTML]{C0C0C0}{\color[HTML]{000000} }} &
  {\color[HTML]{000000} } \\ \midrule
\rowcolor[HTML]{FFFFFF} 
\multicolumn{1}{|c|}{\cellcolor[HTML]{FFFFFF}{\color[HTML]{000000} \textbf{Client status}}} &
  \multicolumn{1}{c|}{\cellcolor[HTML]{FFFFFF}{\color[HTML]{000000} }} &
  \multicolumn{1}{c|}{\cellcolor[HTML]{FFFFFF}{\color[HTML]{000000} \begin{tabular}[c]{@{}c@{}}Client wants\\ to buy:\\ new car\end{tabular}}} &
  \multicolumn{1}{c|}{\cellcolor[HTML]{FFFFFF}{\color[HTML]{000000} }} &
  \multicolumn{1}{c|}{\cellcolor[HTML]{FFFFFF}{\color[HTML]{000000} }} &
  \multicolumn{1}{c|}{\cellcolor[HTML]{FFFFFF}{\color[HTML]{000000} $<$ 0}} &
  \multicolumn{1}{c|}{\cellcolor[HTML]{FFFFFF}{\color[HTML]{000000} }} &
  \multicolumn{1}{c|}{\cellcolor[HTML]{FFFFFF}{\color[HTML]{000000} }} &
  \multicolumn{1}{c|}{\cellcolor[HTML]{FFFFFF}{\color[HTML]{000000} 1 $\leq$ ... $<$ 4 years}} &
  \multicolumn{1}{c|}{\cellcolor[HTML]{FFFFFF}{\color[HTML]{000000} }} &
  \multicolumn{1}{c|}{\cellcolor[HTML]{FFFFFF}{\color[HTML]{000000} }} &
  \multicolumn{1}{c|}{\cellcolor[HTML]{FFFFFF}{\color[HTML]{000000} none registered}} \\ \cmidrule(r){1-1} \cmidrule(l){3-12} 
\rowcolor[HTML]{FFFFFF} 
\multicolumn{1}{|c|}{\cellcolor[HTML]{FFFFFF}{\color[HTML]{000000} \textbf{Recommendation}}} &
  \multicolumn{1}{c|}{\multirow{-4}{*}{\cellcolor[HTML]{FFFFFF}{\color[HTML]{000000} \textbf{\$ 221}}}} &
  \multicolumn{1}{c|}{\cellcolor[HTML]{FFFFFF}{\color[HTML]{000000} used car}} &
  \multicolumn{1}{c|}{\cellcolor[HTML]{FFFFFF}{\color[HTML]{000000} \textbf{}}} &
  \multicolumn{1}{c|}{\cellcolor[HTML]{FFFFFF}{\color[HTML]{000000} \textbf{}}} &
  \multicolumn{1}{c|}{\cellcolor[HTML]{FFFFFF}{\color[HTML]{000000} 0 $\leq$ ... $<$ 200}} &
  \multicolumn{1}{c|}{\cellcolor[HTML]{FFFFFF}{\color[HTML]{000000} }} &
  \multicolumn{1}{c|}{\cellcolor[HTML]{FFFFFF}{\color[HTML]{000000} }} &
  \multicolumn{1}{c|}{\cellcolor[HTML]{FFFFFF}{\color[HTML]{000000} 4 $\leq$ ... $<$ 7 years}} &
  \multicolumn{1}{c|}{\cellcolor[HTML]{FFFFFF}{\color[HTML]{000000} }} &
  \multicolumn{1}{c|}{\cellcolor[HTML]{FFFFFF}{\color[HTML]{000000} }} &
  \multicolumn{1}{c|}{\cellcolor[HTML]{FFFFFF}{\color[HTML]{000000} \begin{tabular}[c]{@{}c@{}}yes, registered under \\ the customer's name\end{tabular}}} \\ \bottomrule
\end{tabular}}
\end{table*}

We see that the obtained diagnoses make sense and provide plausible explanations as to why the credit was denied.

In tab.~\ref{tab:credit_denial_restricted}, we fix the values of a few characteristics on which we suppose the user has no control -- in the example, "Savings account/bonds" and "Present employment since" -- forcing the CF approach to propose changes on the remaining characteristics so that the credit be granted. We also compute the unrestricted CF example for comparison.

\begin{table*}[ht]
\caption{\label{tab:credit_denial_restricted} \small{Decision explanations for three cases of credit denial using restricted CF queries. For each example, the first line "Client status" corresponds to the query point, the second line "Recommendation" corresponds to the unrestricted CF example, and the third line "Restricted recommendation" corresponds to the restricted CF example when the values of the two characteristics Savings account/bonds" and "Present employment since" are fixed.}}
\centering
\resizebox{0.9\textwidth}{!}{\begin{tabular}{@{}ccccccccccccc@{}}
\toprule
\rowcolor[HTML]{FFFFFF} 
\multicolumn{1}{|c|}{\cellcolor[HTML]{FFFFFF}{\color[HTML]{000000} \textbf{}}} &
  \multicolumn{1}{c|}{\cellcolor[HTML]{FFFFFF}{\color[HTML]{000000} \textbf{\begin{tabular}[c]{@{}c@{}}Amount \\ refused\end{tabular}}}} &
  \multicolumn{1}{c|}{\cellcolor[HTML]{FFFFFF}{\color[HTML]{000000} \textbf{Purpose}}} &
  \multicolumn{1}{c|}{\cellcolor[HTML]{FFFFFF}{\color[HTML]{000000} \textbf{\begin{tabular}[c]{@{}c@{}}Credit\\ history\end{tabular}}}} &
  \multicolumn{1}{c|}{\cellcolor[HTML]{FFFFFF}{\color[HTML]{000000} \textbf{\begin{tabular}[c]{@{}c@{}}Installment rate \\ in \% of \\ disposable\\  income\end{tabular}}}} &
  \multicolumn{1}{c|}{\cellcolor[HTML]{FFFFFF}{\color[HTML]{000000} \textbf{\begin{tabular}[c]{@{}c@{}}Status of \\ existing \\ checking \\ account\end{tabular}}}} &
  \multicolumn{1}{c|}{\cellcolor[HTML]{FFFFFF}{\color[HTML]{000000} \textbf{\begin{tabular}[c]{@{}c@{}}Number of \\ existing \\ credits at \\ this bank\end{tabular}}}} &
  \multicolumn{1}{c|}{\cellcolor[HTML]{FFFFFF}{\color[HTML]{000000} \textbf{\begin{tabular}[c]{@{}c@{}}Savings account\\ /bonds\end{tabular}}}} &
  \multicolumn{1}{c|}{\cellcolor[HTML]{FFFFFF}{\color[HTML]{000000} \textbf{\begin{tabular}[c]{@{}c@{}}Present \\ employment\\  since\end{tabular}}}} &
  \multicolumn{1}{c|}{\cellcolor[HTML]{FFFFFF}{\color[HTML]{000000} \textbf{Property}}} &
  \multicolumn{1}{c|}{\cellcolor[HTML]{FFFFFF}{\color[HTML]{000000} \textbf{\begin{tabular}[c]{@{}c@{}}Other debtors /\\  guarantors\end{tabular}}}} &
  \multicolumn{1}{c|}{\cellcolor[HTML]{FFFFFF}{\color[HTML]{000000} \textbf{Telephone}}} &
  \multicolumn{1}{c|}{\cellcolor[HTML]{FFFFFF}{\color[HTML]{000000} \textbf{\begin{tabular}[c]{@{}c@{}}Other \\ installment \\ plans\end{tabular}}}} \\ \midrule
\rowcolor[HTML]{FFFFFF} 
\multicolumn{1}{|c|}{\cellcolor[HTML]{FFFFFF}{\color[HTML]{000000} \textbf{Client status}}} &
  \multicolumn{1}{c|}{\cellcolor[HTML]{FFFFFF}{\color[HTML]{000000} }} &
  \multicolumn{1}{c|}{\cellcolor[HTML]{FFFFFF}{\color[HTML]{000000} }} &
  \multicolumn{1}{c|}{\cellcolor[HTML]{FFFFFF}{\color[HTML]{000000} }} &
  \multicolumn{1}{c|}{\cellcolor[HTML]{FFFFFF}{\color[HTML]{000000} }} &
  \multicolumn{1}{c|}{\cellcolor[HTML]{FFFFFF}{\color[HTML]{000000} $<$ 0}} &
  \multicolumn{1}{c|}{\cellcolor[HTML]{FFFFFF}{\color[HTML]{000000} }} &
  \multicolumn{1}{c|}{\cellcolor[HTML]{FFFFFF}{\color[HTML]{000000} $<$  100}} &
  \multicolumn{1}{c|}{\cellcolor[HTML]{FFFFFF}{\color[HTML]{000000} }} &
  \multicolumn{1}{c|}{\cellcolor[HTML]{FFFFFF}{\color[HTML]{000000} }} &
  \multicolumn{1}{c|}{\cellcolor[HTML]{FFFFFF}{\color[HTML]{000000} }} &
  \multicolumn{1}{c|}{\cellcolor[HTML]{FFFFFF}{\color[HTML]{000000} }} &
  \multicolumn{1}{c|}{\cellcolor[HTML]{FFFFFF}{\color[HTML]{000000} stores}} \\ \cmidrule(r){1-1} \cmidrule(l){3-13} 
\rowcolor[HTML]{FFFFFF} 
\multicolumn{1}{|c|}{\cellcolor[HTML]{FFFFFF}{\color[HTML]{000000} \textbf{Recommendation}}} &
  \multicolumn{1}{c|}{\cellcolor[HTML]{FFFFFF}{\color[HTML]{000000} }} &
  \multicolumn{1}{c|}{\cellcolor[HTML]{FFFFFF}{\color[HTML]{000000} }} &
  \multicolumn{1}{c|}{\cellcolor[HTML]{FFFFFF}{\color[HTML]{000000} }} &
  \multicolumn{1}{c|}{\cellcolor[HTML]{FFFFFF}{\color[HTML]{000000} }} &
  \multicolumn{1}{c|}{\cellcolor[HTML]{FFFFFF}{\color[HTML]{000000} }} &
  \multicolumn{1}{c|}{\cellcolor[HTML]{FFFFFF}{\color[HTML]{000000} }} &
  \multicolumn{1}{c|}{\cellcolor[HTML]{FFFFFF}{\color[HTML]{000000} 500 $\leq$ ... $<$  1000}} &
  \multicolumn{1}{c|}{\cellcolor[HTML]{FFFFFF}{\color[HTML]{000000} }} &
  \multicolumn{1}{c|}{\cellcolor[HTML]{FFFFFF}{\color[HTML]{000000} }} &
  \multicolumn{1}{c|}{\cellcolor[HTML]{FFFFFF}{\color[HTML]{000000} }} &
  \multicolumn{1}{c|}{\cellcolor[HTML]{FFFFFF}{\color[HTML]{000000} }} &
  \multicolumn{1}{c|}{\cellcolor[HTML]{FFFFFF}{\color[HTML]{000000} }} \\ \cmidrule(r){1-1} \cmidrule(l){3-13} 
\rowcolor[HTML]{FFFFFF} 
\multicolumn{1}{|c|}{\cellcolor[HTML]{FFFFFF}{\color[HTML]{000000} \textbf{\begin{tabular}[c]{@{}c@{}}Restricted\\ recommendation\end{tabular}}}} &
  \multicolumn{1}{c|}{\multirow{-3}{*}{\cellcolor[HTML]{FFFFFF}{\color[HTML]{000000} \textbf{\$ 778}}}} &
  \multicolumn{1}{c|}{\cellcolor[HTML]{FFFFFF}{\color[HTML]{000000} \textbf{}}} &
  \multicolumn{1}{c|}{\cellcolor[HTML]{FFFFFF}{\color[HTML]{000000} \textbf{}}} &
  \multicolumn{1}{c|}{\cellcolor[HTML]{FFFFFF}{\color[HTML]{000000} }} &
  \multicolumn{1}{c|}{\cellcolor[HTML]{FFFFFF}{\color[HTML]{000000} $\geq$ 200}} &
  \multicolumn{1}{c|}{\cellcolor[HTML]{FFFFFF}{\color[HTML]{000000} }} &
  \multicolumn{1}{c|}{\cellcolor[HTML]{FFFFFF}{\color[HTML]{000000} }} &
  \multicolumn{1}{c|}{\cellcolor[HTML]{FFFFFF}{\color[HTML]{000000} }} &
  \multicolumn{1}{c|}{\cellcolor[HTML]{FFFFFF}{\color[HTML]{000000} }} &
  \multicolumn{1}{c|}{\cellcolor[HTML]{FFFFFF}{\color[HTML]{000000} }} &
  \multicolumn{1}{c|}{\cellcolor[HTML]{FFFFFF}{\color[HTML]{000000} }} &
  \multicolumn{1}{c|}{\cellcolor[HTML]{FFFFFF}{\color[HTML]{000000} none}} \\ \midrule
\rowcolor[HTML]{C0C0C0} 
\multicolumn{1}{l}{\cellcolor[HTML]{C0C0C0}{\color[HTML]{000000} }} &
  \multicolumn{1}{l}{\cellcolor[HTML]{C0C0C0}{\color[HTML]{000000} }} &
  \multicolumn{1}{l}{\cellcolor[HTML]{C0C0C0}{\color[HTML]{000000} }} &
  \multicolumn{1}{l}{\cellcolor[HTML]{C0C0C0}{\color[HTML]{000000} }} &
  \multicolumn{1}{l}{\cellcolor[HTML]{C0C0C0}{\color[HTML]{000000} }} &
  \multicolumn{1}{l}{\cellcolor[HTML]{C0C0C0}{\color[HTML]{000000} }} &
  \multicolumn{1}{l}{\cellcolor[HTML]{C0C0C0}{\color[HTML]{000000} }} &
  \multicolumn{1}{l}{\cellcolor[HTML]{C0C0C0}{\color[HTML]{000000} }} &
  \multicolumn{1}{l}{\cellcolor[HTML]{C0C0C0}{\color[HTML]{000000} }} &
  \multicolumn{1}{l}{\cellcolor[HTML]{C0C0C0}{\color[HTML]{000000} }} &
  \multicolumn{1}{l}{\cellcolor[HTML]{C0C0C0}{\color[HTML]{000000} }} &
  \multicolumn{1}{l}{\cellcolor[HTML]{C0C0C0}{\color[HTML]{000000} }} &
  \multicolumn{1}{l}{\cellcolor[HTML]{C0C0C0}{\color[HTML]{000000} }} \\ \midrule
\rowcolor[HTML]{FFFFFF} 
\multicolumn{1}{|c|}{\cellcolor[HTML]{FFFFFF}{\color[HTML]{000000} \textbf{Client status}}} &
  \multicolumn{1}{c|}{\cellcolor[HTML]{FFFFFF}{\color[HTML]{000000} }} &
  \multicolumn{1}{c|}{\cellcolor[HTML]{FFFFFF}{\color[HTML]{000000} car new}} &
  \multicolumn{1}{c|}{\cellcolor[HTML]{FFFFFF}{\color[HTML]{000000} }} &
  \multicolumn{1}{c|}{\cellcolor[HTML]{FFFFFF}{\color[HTML]{000000} }} &
  \multicolumn{1}{c|}{\cellcolor[HTML]{FFFFFF}{\color[HTML]{000000} }} &
  \multicolumn{1}{c|}{\cellcolor[HTML]{FFFFFF}{\color[HTML]{000000} }} &
  \multicolumn{1}{c|}{\cellcolor[HTML]{FFFFFF}{\color[HTML]{000000} $<$  100}} &
  \multicolumn{1}{c|}{\cellcolor[HTML]{FFFFFF}{\color[HTML]{000000} }} &
  \multicolumn{1}{c|}{\cellcolor[HTML]{FFFFFF}{\color[HTML]{000000} }} &
  \multicolumn{1}{c|}{\cellcolor[HTML]{FFFFFF}{\color[HTML]{000000} }} &
  \multicolumn{1}{c|}{\cellcolor[HTML]{FFFFFF}{\color[HTML]{000000} }} &
  \multicolumn{1}{c|}{\cellcolor[HTML]{FFFFFF}{\color[HTML]{000000} }} \\ \cmidrule(r){1-1} \cmidrule(l){3-13} 
\rowcolor[HTML]{FFFFFF} 
\multicolumn{1}{|c|}{\cellcolor[HTML]{FFFFFF}{\color[HTML]{000000} \textbf{Recommendation}}} &
  \multicolumn{1}{c|}{\cellcolor[HTML]{FFFFFF}{\color[HTML]{000000} }} &
  \multicolumn{1}{c|}{\cellcolor[HTML]{FFFFFF}{\color[HTML]{000000} \textbf{}}} &
  \multicolumn{1}{c|}{\cellcolor[HTML]{FFFFFF}{\color[HTML]{000000} }} &
  \multicolumn{1}{c|}{\cellcolor[HTML]{FFFFFF}{\color[HTML]{000000} }} &
  \multicolumn{1}{c|}{\cellcolor[HTML]{FFFFFF}{\color[HTML]{000000} }} &
  \multicolumn{1}{c|}{\cellcolor[HTML]{FFFFFF}{\color[HTML]{000000} }} &
  \multicolumn{1}{c|}{\cellcolor[HTML]{FFFFFF}{\color[HTML]{000000} 500 $\leq$ ... $<$  1000}} &
  \multicolumn{1}{c|}{\cellcolor[HTML]{FFFFFF}{\color[HTML]{000000} }} &
  \multicolumn{1}{c|}{\cellcolor[HTML]{FFFFFF}{\color[HTML]{000000} }} &
  \multicolumn{1}{c|}{\cellcolor[HTML]{FFFFFF}{\color[HTML]{000000} }} &
  \multicolumn{1}{c|}{\cellcolor[HTML]{FFFFFF}{\color[HTML]{000000} }} &
  \multicolumn{1}{c|}{\cellcolor[HTML]{FFFFFF}{\color[HTML]{000000} }} \\ \cmidrule(r){1-1} \cmidrule(l){3-13} 
\rowcolor[HTML]{FFFFFF} 
\multicolumn{1}{|c|}{\cellcolor[HTML]{FFFFFF}{\color[HTML]{000000} \textbf{\begin{tabular}[c]{@{}c@{}}Restricted\\ recommendation\end{tabular}}}} &
  \multicolumn{1}{c|}{\multirow{-3}{*}{\cellcolor[HTML]{FFFFFF}{\color[HTML]{000000} \textbf{\$ 797}}}} &
  \multicolumn{1}{c|}{\cellcolor[HTML]{FFFFFF}{\color[HTML]{000000} car used}} &
  \multicolumn{1}{c|}{\cellcolor[HTML]{FFFFFF}{\color[HTML]{000000} \textbf{}}} &
  \multicolumn{1}{c|}{\cellcolor[HTML]{FFFFFF}{\color[HTML]{000000} \textbf{}}} &
  \multicolumn{1}{c|}{\cellcolor[HTML]{FFFFFF}{\color[HTML]{000000} }} &
  \multicolumn{1}{c|}{\cellcolor[HTML]{FFFFFF}{\color[HTML]{000000} }} &
  \multicolumn{1}{c|}{\cellcolor[HTML]{FFFFFF}{\color[HTML]{000000} }} &
  \multicolumn{1}{c|}{\cellcolor[HTML]{FFFFFF}{\color[HTML]{000000} }} &
  \multicolumn{1}{c|}{\cellcolor[HTML]{FFFFFF}{\color[HTML]{000000} }} &
  \multicolumn{1}{c|}{\cellcolor[HTML]{FFFFFF}{\color[HTML]{000000} }} &
  \multicolumn{1}{c|}{\cellcolor[HTML]{FFFFFF}{\color[HTML]{000000} }} &
  \multicolumn{1}{c|}{\cellcolor[HTML]{FFFFFF}{\color[HTML]{000000} }} \\ \midrule
\rowcolor[HTML]{C0C0C0} 
\multicolumn{1}{l}{\cellcolor[HTML]{C0C0C0}{\color[HTML]{000000} }} &
  \multicolumn{1}{l}{\cellcolor[HTML]{C0C0C0}{\color[HTML]{000000} }} &
  \multicolumn{1}{l}{\cellcolor[HTML]{C0C0C0}{\color[HTML]{000000} }} &
  \multicolumn{1}{l}{\cellcolor[HTML]{C0C0C0}{\color[HTML]{000000} }} &
  \multicolumn{1}{l}{\cellcolor[HTML]{C0C0C0}{\color[HTML]{000000} }} &
  \multicolumn{1}{l}{\cellcolor[HTML]{C0C0C0}{\color[HTML]{000000} }} &
  \multicolumn{1}{l}{\cellcolor[HTML]{C0C0C0}{\color[HTML]{000000} }} &
  \multicolumn{1}{l}{\cellcolor[HTML]{C0C0C0}{\color[HTML]{000000} }} &
  \multicolumn{1}{l}{\cellcolor[HTML]{C0C0C0}{\color[HTML]{000000} }} &
  \multicolumn{1}{l}{\cellcolor[HTML]{C0C0C0}{\color[HTML]{000000} }} &
  \multicolumn{1}{l}{\cellcolor[HTML]{C0C0C0}{\color[HTML]{000000} }} &
  \multicolumn{1}{l}{\cellcolor[HTML]{C0C0C0}{\color[HTML]{000000} }} &
  \multicolumn{1}{l}{\cellcolor[HTML]{C0C0C0}{\color[HTML]{000000} }} \\ \midrule
\rowcolor[HTML]{FFFFFF} 
\multicolumn{1}{|c|}{\cellcolor[HTML]{FFFFFF}{\color[HTML]{000000} \textbf{Client status}}} &
  \multicolumn{1}{c|}{\cellcolor[HTML]{FFFFFF}{\color[HTML]{000000} }} &
  \multicolumn{1}{c|}{\cellcolor[HTML]{FFFFFF}{\color[HTML]{000000} }} &
  \multicolumn{1}{c|}{\cellcolor[HTML]{FFFFFF}{\color[HTML]{000000} }} &
  \multicolumn{1}{c|}{\cellcolor[HTML]{FFFFFF}{\color[HTML]{000000} }} &
  \multicolumn{1}{c|}{\cellcolor[HTML]{FFFFFF}{\color[HTML]{000000} }} &
  \multicolumn{1}{c|}{\cellcolor[HTML]{FFFFFF}{\color[HTML]{000000} 0 $\geq$ ... $<$ 200}} &
  \multicolumn{1}{c|}{\cellcolor[HTML]{FFFFFF}{\color[HTML]{000000} }} &
  \multicolumn{1}{c|}{\cellcolor[HTML]{FFFFFF}{\color[HTML]{000000} 1 $\leq$ ... $<$ 4 years}} &
  \multicolumn{1}{c|}{\cellcolor[HTML]{FFFFFF}{\color[HTML]{000000} Real estate}} &
  \multicolumn{1}{c|}{\cellcolor[HTML]{FFFFFF}{\color[HTML]{000000} }} &
  \multicolumn{1}{c|}{\cellcolor[HTML]{FFFFFF}{\color[HTML]{000000} }} &
  \multicolumn{1}{c|}{\cellcolor[HTML]{FFFFFF}{\color[HTML]{000000} }} \\ \cmidrule(r){1-1} \cmidrule(l){3-13} 
\rowcolor[HTML]{FFFFFF} 
\multicolumn{1}{|c|}{\cellcolor[HTML]{FFFFFF}{\color[HTML]{000000} \textbf{Recommendation}}} &
  \multicolumn{1}{c|}{\cellcolor[HTML]{FFFFFF}{\color[HTML]{000000} }} &
  \multicolumn{1}{c|}{\cellcolor[HTML]{FFFFFF}{\color[HTML]{000000} \textbf{}}} &
  \multicolumn{1}{c|}{\cellcolor[HTML]{FFFFFF}{\color[HTML]{000000} \textbf{}}} &
  \multicolumn{1}{c|}{\cellcolor[HTML]{FFFFFF}{\color[HTML]{000000} }} &
  \multicolumn{1}{c|}{\cellcolor[HTML]{FFFFFF}{\color[HTML]{000000} }} &
  \multicolumn{1}{c|}{\cellcolor[HTML]{FFFFFF}{\color[HTML]{000000} $\geq$ 200}} &
  \multicolumn{1}{c|}{\cellcolor[HTML]{FFFFFF}{\color[HTML]{000000} }} &
  \multicolumn{1}{c|}{\cellcolor[HTML]{FFFFFF}{\color[HTML]{000000} 4 $\leq$ ... $<$ 7 years}} &
  \multicolumn{1}{c|}{\cellcolor[HTML]{FFFFFF}{\color[HTML]{000000} }} &
  \multicolumn{1}{c|}{\cellcolor[HTML]{FFFFFF}{\color[HTML]{000000} }} &
  \multicolumn{1}{c|}{\cellcolor[HTML]{FFFFFF}{\color[HTML]{000000} }} &
  \multicolumn{1}{c|}{\cellcolor[HTML]{FFFFFF}{\color[HTML]{000000} }} \\ \cmidrule(r){1-1} \cmidrule(l){3-13} 
\rowcolor[HTML]{FFFFFF} 
\multicolumn{1}{|c|}{\cellcolor[HTML]{FFFFFF}{\color[HTML]{000000} \textbf{\begin{tabular}[c]{@{}c@{}}Restricted\\ recommendation\end{tabular}}}} &
  \multicolumn{1}{c|}{\multirow{-3}{*}{\cellcolor[HTML]{FFFFFF}{\color[HTML]{000000} \textbf{\$ 707}}}} &
  \multicolumn{1}{c|}{\cellcolor[HTML]{FFFFFF}{\color[HTML]{000000} \textbf{}}} &
  \multicolumn{1}{c|}{\cellcolor[HTML]{FFFFFF}{\color[HTML]{000000} }} &
  \multicolumn{1}{c|}{\cellcolor[HTML]{FFFFFF}{\color[HTML]{000000} \textbf{}}} &
  \multicolumn{1}{c|}{\cellcolor[HTML]{FFFFFF}{\color[HTML]{000000} \textbf{}}} &
  \multicolumn{1}{c|}{\cellcolor[HTML]{FFFFFF}{\color[HTML]{000000} $\geq$ 200}} &
  \multicolumn{1}{c|}{\cellcolor[HTML]{FFFFFF}{\color[HTML]{000000} }} &
  \multicolumn{1}{c|}{\cellcolor[HTML]{FFFFFF}{\color[HTML]{000000} }} &
  \multicolumn{1}{c|}{\cellcolor[HTML]{FFFFFF}{\color[HTML]{000000} \begin{tabular}[c]{@{}c@{}}building society \\ savings agreement/ \\ life insurance\end{tabular}}} &
  \multicolumn{1}{c|}{\cellcolor[HTML]{FFFFFF}{\color[HTML]{000000} }} &
  \multicolumn{1}{c|}{\cellcolor[HTML]{FFFFFF}{\color[HTML]{000000} }} &
  \multicolumn{1}{c|}{\cellcolor[HTML]{FFFFFF}{\color[HTML]{000000} }} \\ \bottomrule
\end{tabular}}
\end{table*}

\subsection{Use case 2}

In this use case, we consider binary classification problems extracted from the MNIST digits dataset. We study in particular classes that are naturally ambiguous (such as 5 and 6 for instance). We use the CF approach of the paper to emphasize the parts of a miss-classified image that led the classifier into miss-classifying this input image. We consider that the miss-classification is due to the input image being ambiguous, and not to the classifier being bad at classifying certain images. The reason for the miss-classification is thus exclusively sought for in the input data (image pixels), and not in the classifier learned parameters/decision regions (this would amount then to perform model debugging, a completely different problem not addressed in this paper).

We do not use the raw pixel data directly, but the projections of the images in a PCA basis composed of the first 50 principal components, which is enough to obtain good quality reconstructions of the original images. As a classifier, we use a XGBoost binary classification model consisting of 200 trees of maximal depth 7.

\subsubsection{Visual assessement}
In this use case, we outline visually the changes in the CF examples by comparing them to their respective query images. The latter are chosen among miss-classified images for which even the human eye cannot really tell to which class they belong. 
The changes can be made more obvious by varying the binary decision threshold of the classification model. In the paper, we have presented a general implementation of the CF approach for multi-class classification. For binary classification problems, XGBoost and LightGBM  use the same model configuration and training criteria as for binary logistic regression. The output of the model is a 1-dimensional score constrained to the interval $[0, 1]$ by using a logistic aggregation function as detailed in sec.~\ref{sec:geom_carac}. For binary classification, this score represents the probability of the predicted element to belong to the first class, and, as such, the class prediction is obtained by comparing this score to a value $\epsilon$ (equal to $0.5$ by default). The binary CF approach is thus equivalent to solving the problem:
\begin{align} \label{eq:request_binary}
\textnormal{CF}(X, j) = \arg \min_{\left\{Y \left| F(Y) \leq \epsilon \right.\right\}} \norm{Y - X}_2^2
\end{align}

\begin{figure*}
    \centering
    \includegraphics[width=0.8\textwidth]{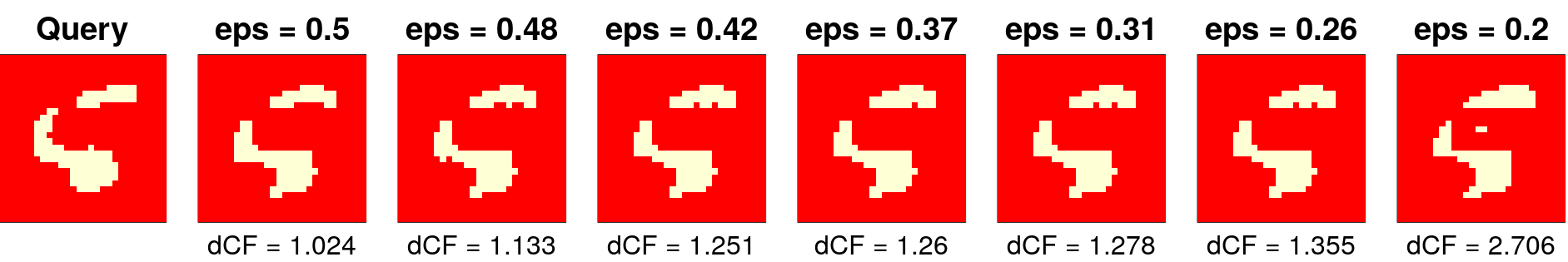}
    \caption{\small{Example of a "5" miss-classified as a "6" by the model. The initial query image is represented on the left. We make vary the decision threshold $\epsilon$ between 0.5 and 0.2. Getting a CF example that is classified into the right class with more confidence by the model is done at the expense of the distortion introduced to the initial query data, i.e., the lower the decision threshold "eps", the higher the distance to the initial query "dCF". Visually, we see that the CF approach provides plausible changes of the initial query data to make it look more like a "5".}}
    \label{fig:CF_6_5}
\end{figure*}

\begin{figure*}
    \centering
    \includegraphics[width=0.8\textwidth]{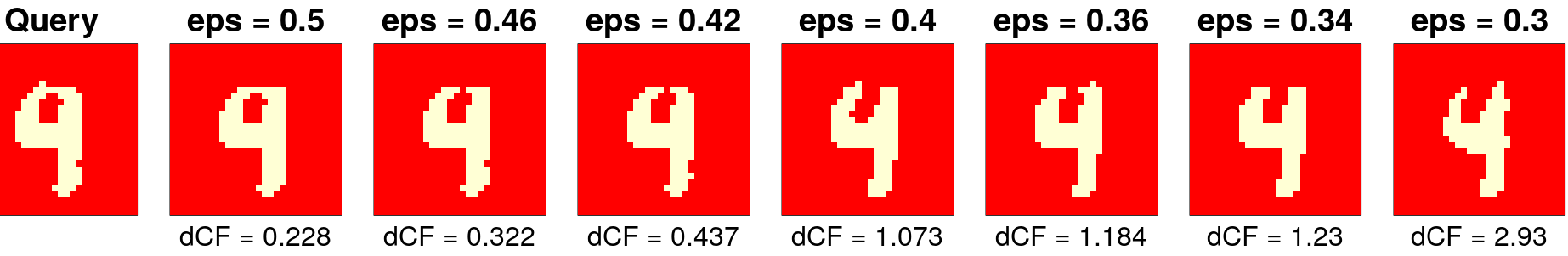}
    \caption{\small{Example of a "4" miss-classified as a "9" by the model. We represent the CF examples associated with decision thresholds ("eps") varying between 0.5 and 0.3. Each time, we report the distance to the initial query image ("dCF").}}
    \label{fig:CF_9_4}
\end{figure*}

Varying the threshold $\epsilon$ for a given miss-classified query image $X$ causes the CF approach to find CF examples that are classified in the right class with more and more confidence by the model. The sequences of images in the fig.~\ref{fig:CF_6_5} and \ref{fig:CF_9_4} represents the PCA reconstructions of the CF examples for decreasing values of $\epsilon$ (i.e., given a two-class classification problem involving two of the ten classes of MNIST, we consider images whose true class is the first class, but which are miss-classified in the second one: decreasing $\epsilon$ produces CF examples derived from the query that are classified in the first class with more and more confidence). We notice that, as the decision threshold decreases, the CF examples become also less and less ambiguous perceptually speaking.

\subsubsection{Numerical assessment}

We assess the quality of the CF examples provided in terms of three numerical criteria: stability, sparsity and distortion (L1 / L2 distance between the query point and the corresponding CF example). The sparsity is computed as the number of characteristics which get modified in the CF example compared to the original query. The stability is computed using the eq.~2 of \cite{alvarez2018robustness}, taking as $f(x)$ the function which associates a point $x$ to its closest CF example.

We compare ourselves to the CF approach described in \cite{lucic2019focus}, and referred to as FOCUS. This approach has the advantage of being specific to tree ensemble models, and does not suffer from scalability problems on large models. We re-implemented the approach so that it can be used with XGBoost models. We also verify that the produced CF example is valid with respect to the initial model (and not only with respect to the differentiable tree ensemble surrogate used by the method). 

All the CF examples in the examples below are computed for a confidence threshold $\epsilon$ equal to 0.5. In the tab.~\ref{tab:crit_num} below, we show the comparison between the FOCUS method and our approach for 4 different binary classification problems: "5" versus "6", "1" versus "7", "4" versus "9", and "3" versus "8". For each miss-classified example, we compute the associated CF example, and the induced sparsity, distortion, and stability. For a given binary classification problem, the results are averaged over all the miss-classified examples associated with this problem.

\begin{table}[H]
\caption{\label{tab:crit_num} \small{Assessing the average sparsity, distortion, and stability over all miss-classified examples for four binary classification problems. The best scores are stressed in bold.}}
\centering
\resizebox{0.4\textwidth}{!}{\begin{tabular}{@{}cccccc@{}}
\toprule
\multicolumn{1}{|c|}{{\color[HTML]{000000} \textbf{}}} &
  \multicolumn{1}{c|}{\cellcolor[HTML]{FFFFFF}{\color[HTML]{000000} \textbf{problem}}} &
  \multicolumn{1}{c|}{\cellcolor[HTML]{FFFFFF}{\color[HTML]{000000} \textbf{L1 distortion}}} &
  \multicolumn{1}{c|}{\cellcolor[HTML]{FFFFFF}{\color[HTML]{000000} \textbf{L2 distortion}}} &
  \multicolumn{1}{c|}{\cellcolor[HTML]{FFFFFF}{\color[HTML]{000000} \textbf{sparsity}}} &
  \multicolumn{1}{c|}{\cellcolor[HTML]{FFFFFF}{\color[HTML]{000000} \textbf{stability}}} \\ \midrule
\multicolumn{1}{|c|}{{\color[HTML]{000000} \textbf{FOCUS}}} &
  \multicolumn{1}{c|}{{\color[HTML]{000000} }} &
  \multicolumn{1}{c|}{{\color[HTML]{000000} 0.93}} &
  \multicolumn{1}{c|}{{\color[HTML]{000000} 0.37}} &
  \multicolumn{1}{c|}{{\color[HTML]{000000} 15.83}} &
  \multicolumn{1}{c|}{{\color[HTML]{000000} 1.07}} \\ \cmidrule(r){1-1} \cmidrule(l){3-6} 
\multicolumn{1}{|c|}{{\color[HTML]{000000} \textbf{Ours}}} &
  \multicolumn{1}{c|}{\multirow{-2}{*}{{\color[HTML]{000000} 5 versus 6}}} &
  \multicolumn{1}{c|}{{\color[HTML]{000000} \bf{0.32}}} &
  \multicolumn{1}{c|}{{\color[HTML]{000000} \bf{0.2}}} &
  \multicolumn{1}{c|}{{\color[HTML]{000000} \bf{3.08}}} &
  \multicolumn{1}{c|}{{\color[HTML]{000000} \bf{0.98}}} \\ \midrule
\rowcolor[HTML]{C0C0C0} 
{\color[HTML]{000000} } &
  {\color[HTML]{000000} } &
  {\color[HTML]{000000} } &
  {\color[HTML]{000000} } &
  {\color[HTML]{000000} } &
  {\color[HTML]{000000} } \\ \midrule
\multicolumn{1}{|c|}{{\color[HTML]{000000} \textbf{FOCUS}}} &
  \multicolumn{1}{c|}{{\color[HTML]{000000} }} &
  \multicolumn{1}{c|}{{\color[HTML]{000000} 0.66}} &
  \multicolumn{1}{c|}{{\color[HTML]{000000} 0.28}} &
  \multicolumn{1}{c|}{{\color[HTML]{000000} 10.75}} &
  \multicolumn{1}{c|}{{\color[HTML]{000000} 1.04}} \\ \cmidrule(r){1-1} \cmidrule(l){3-6} 
\multicolumn{1}{|c|}{{\color[HTML]{000000} \textbf{Ours}}} &
  \multicolumn{1}{c|}{\multirow{-2}{*}{{\color[HTML]{000000} 1 versus 7}}} &
  \multicolumn{1}{c|}{{\color[HTML]{000000} \bf{0.17}}} &
  \multicolumn{1}{c|}{{\color[HTML]{000000} \bf{0.14}}} &
  \multicolumn{1}{c|}{{\color[HTML]{000000} \bf{1.75}}} &
  \multicolumn{1}{c|}{{\color[HTML]{000000} \bf{0.96}}} \\ \midrule
\rowcolor[HTML]{C0C0C0} 
{\color[HTML]{000000} } &
  {\color[HTML]{000000} } &
  {\color[HTML]{000000} } &
  {\color[HTML]{000000} } &
  {\color[HTML]{000000} } &
  {\color[HTML]{000000} } \\ \midrule
\multicolumn{1}{|c|}{{\color[HTML]{000000} \textbf{FOCUS}}} &
  \multicolumn{1}{c|}{{\color[HTML]{000000} }} &
  \multicolumn{1}{c|}{{\color[HTML]{000000} 1.04}} &
  \multicolumn{1}{c|}{{\color[HTML]{000000} 0.31}} &
  \multicolumn{1}{c|}{{\color[HTML]{000000} 20.91}} &
  \multicolumn{1}{c|}{{\color[HTML]{000000} 1.22}} \\ \cmidrule(r){1-1} \cmidrule(l){3-6} 
\multicolumn{1}{|c|}{{\color[HTML]{000000} \textbf{Ours}}} &
  \multicolumn{1}{c|}{\multirow{-2}{*}{{\color[HTML]{000000} 4 versus 9}}} &
  \multicolumn{1}{c|}{{\color[HTML]{000000} \bf{0.26}}} &
  \multicolumn{1}{c|}{{\color[HTML]{000000} \bf{0.15}}} &
  \multicolumn{1}{c|}{{\color[HTML]{000000} \bf{3.36}}} &
  \multicolumn{1}{c|}{{\color[HTML]{000000} \bf{0.9}}} \\ \midrule
\rowcolor[HTML]{C0C0C0} 
{\color[HTML]{000000} } &
  {\color[HTML]{000000} } &
  {\color[HTML]{000000} } &
  {\color[HTML]{000000} } &
  {\color[HTML]{000000} } &
  {\color[HTML]{000000} } \\ \midrule
\multicolumn{1}{|c|}{{\color[HTML]{000000} \textbf{FOCUS}}} &
  \multicolumn{1}{c|}{{\color[HTML]{000000} }} &
  \multicolumn{1}{c|}{{\color[HTML]{000000} 1.07}} &
  \multicolumn{1}{c|}{{\color[HTML]{000000} 0.38}} &
  \multicolumn{1}{c|}{{\color[HTML]{000000} 18.5}} &
  \multicolumn{1}{c|}{{\color[HTML]{000000} 1.04}} \\ \cmidrule(r){1-1} \cmidrule(l){3-6} 
\multicolumn{1}{|c|}{{\color[HTML]{000000} \textbf{Ours}}} &
  \multicolumn{1}{c|}{\multirow{-2}{*}{{\color[HTML]{000000} 3 versus 8}}} &
  \multicolumn{1}{c|}{{\color[HTML]{000000} \bf{0.29}}} &
  \multicolumn{1}{c|}{{\color[HTML]{000000} \bf{0.2}}} &
  \multicolumn{1}{c|}{{\color[HTML]{000000} \bf{3}}} &
  \multicolumn{1}{c|}{{\color[HTML]{000000} \bf{0.94}}} \\ \bottomrule
\end{tabular}}
\end{table}

\subsection{Use case 3}

In this use case, we propose an adaptation of the CF approach described in the paper to perform counterfactual reasoning on regression problems.

Applying the method of the paper to regression requires no special modification of the algorithm. The aggregation function in the eq.~\ref{eq:aggr} is replaced by the identity (unless we are performing logistic regression, in which case, it remains a sigmoid). Concretely, in the algo.~\ref{algo:request}, we replace the instruction line 35 testing if the class of the decision region being analyzed is the class targeted for the CF example by a test to see if the prediction of the model in that region lies within the targeted interval for the CF example. This amounts to solving the problem:
\begin{align} \label{eq:regression_query}
\textnormal{CF}(X, j) = \arg \min_{\left\{Y \left| F(Y) \in \Omega_\textnormal{target} \right.\right\}} \norm{Y - X}_2^2
\end{align}
where $F$ is the regression model and $\Omega_\textnormal{target}$ is the target domain where the prediction of the model should lie. In the experiments, we set $\Omega_\textnormal{target}$ to be an interval.

We use a dataset describing the sale of individual residential property in Ames,
Iowa from 2006 to 2010. The data set contains 2930 observations and a large number of
explanatory variables (23 nominal, 23 ordinal, 14 discrete, and 20 continuous) involved in
assessing home values \cite{de2011ames}\footnote{https://www.kaggle.com/c/house-prices-advanced-regression-techniques/}. These variables describe the characteristics of the house, as well as that of the neighborhood.
We apply our algorithm to a XGBoost regression model trained to predict the sale price of a house given a set of explanatory variables, and we try to answer three types of questions:
\begin{itemize}
\item Case 1: point of view of the seller: what would be the changes/repairs to perform a minima in the house to increase the sale price to a target price ? We try to answer this question by considering only the variables that it is possible to change. For instance, it is not possible to change variables such as the total area, the date of construction, the quality of the neighborhood ... So, we keep these variables fixed, applying the CF approach to the remaining set of variables.
\item Case 2: point of view of the buyer: what would be the characteristics on which a buyer would have to make some concessions to have a chance to find a house whose price is lower by x-percents ? We also include the possibility for the buyer to fix the characteristics on which he is not ready to make any concession.
\item Case 3: point of view of the buyer: given a fixed budget, what choices do the buyer have within that budget in terms of several characteristics of interest defined by the buyer ?
\end{itemize}

All variables are re-scaled between 0 and 1 to avoid one variable having more weight in the CF example computation. Categorical variables that do not possess a natural order are reordered according to a subjective order. For instance, building materials are ordered according to their price and/or installation cost. Categorical indexes are re-scaled between 0 and 1, 0 referring to the worst appreciation, and 1 and to the best one according to the subjective/natural order defined.

\subsubsection{Case 1}

The results for this use case are shown in the tab.~\ref{tab:house_1}, \ref{tab:house_2}, and \ref{tab:house_2_bis}. They represent the minimal changes to perform in a house so that it could be sold within a higher price range. Some characteristics which cannot be modified such as the location, the nature of the neighborhood, or the overall lot area are fixed. The computation of the CF example is thus restricted to the characteristics which can be changed such as the interior configuration, the finish and/or the materials rating in specific parts of the house such as the kitchen ... We perform a restricted CF query on these characteristics in the sense of def.~\ref{eq:restricted_query}.

\begin{table}[H]
\centering
\resizebox{\linewidth}{!}{
\begin{tabular}[t]{>{}lllllllll}
\toprule
\textbf{ } & \textbf{OverallCond} & \textbf{RoofMatl} & \textbf{BsmtFinSF1} & \textbf{KitchenQual} & \textbf{GarageType} & \textbf{WoodDeckSF} & \textbf{OpenPorchSF} & \textbf{ScreenPorch}\\
\midrule
\cellcolor{gray!6}{\textbf{initial}} & \cellcolor{gray!6}{5} & \cellcolor{gray!6}{CompShg} & \cellcolor{gray!6}{0} & \cellcolor{gray!6}{Gd} & \cellcolor{gray!6}{BuiltIn} & \cellcolor{gray!6}{0} & \cellcolor{gray!6}{36} & \cellcolor{gray!6}{0}\\
\textbf{CF} & 6 & Membran & 811 & Ex & Basement & 26 & 95 & 101\\
\bottomrule
\end{tabular}}
\caption{\label{tab:house_1} \small{House with an initial sale price of \$ 168810. Minimal changes to perform to sell it within a target interval [\$ 199010, \$ 250000]. The changes suggested are among others to improve the house overall condition, to change the roof material, to increase the finished area of the basement (BsmtFinSF1), to improve the kitchen quality rating (KitchenQual), to change the type of garage from "built-in" to "basement" type, to add a screen porch of 101 SF (ScreenPorch) ... New sale price estimation: \$ 199014.}}
\end{table}

\begin{table}[H]
\centering
\resizebox{\linewidth}{!}{
\begin{tabular}[t]{>{}llllll}
\toprule
\textbf{ } & \textbf{RoofMatl} & \textbf{BsmtFinSF1} & \textbf{BsmtUnfSF} & \textbf{OpenPorchSF} & \textbf{ScreenPorch}\\
\midrule
\cellcolor{gray!6}{\textbf{initial}} & \cellcolor{gray!6}{CompShg} & \cellcolor{gray!6}{0} & \cellcolor{gray!6}{585} & \cellcolor{gray!6}{112} & \cellcolor{gray!6}{147}\\
\textbf{CF} & Membran & 601.5 & 474.98 & 113.5 & 160.5\\
\bottomrule
\end{tabular}}
\caption{\label{tab:house_2} \small{House with an initial sale price of \$ 195832. Minimal changes to perform to sell it within a target interval [\$ 203382, \$ 250000]. The changes suggested are among others to change the roof material (RoofMatl), to decrease the number of unfinished square feet in the basement (BsmtUnfSF) ... New sale price estimation: \$ 203474.}}
\end{table}

\begin{table}[H]
\centering
\resizebox{\linewidth}{!}{
\begin{tabular}[t]{>{}llllll}
\toprule
\textbf{ } & \textbf{RoofMatl} & \textbf{KitchenQual} & \textbf{Fireplaces} & \textbf{OpenPorchSF} & \textbf{ScreenPorch}\\
\midrule
\cellcolor{gray!6}{\textbf{initial}} & \cellcolor{gray!6}{CompShg} & \cellcolor{gray!6}{TA} & \cellcolor{gray!6}{1} & \cellcolor{gray!6}{112} & \cellcolor{gray!6}{147}\\
\textbf{CF} & Membran & Ex & 2 & 168 & 191\\
\bottomrule
\end{tabular}}
\caption{\label{tab:house_2_bis} \small{Minimal changes to perform to sell the same house as in tab.~\ref{tab:house_2}, this time within a target interval [\$ 211442, \$ 250000]. New sale price estimation: \$ 211496.}}
\end{table}

\subsubsection{Case 2}

The results for this use case are shown in the tab.~\ref{tab:buyer1}, and tab.~\ref{tab:buyer2}. We suppose that the buyer selects a house in the catalog which corresponds to an ideal house for him. We use this house as the query point to find the CF example lying in the target interval specified. In this use case, the buyer wants to find a cheaper house, so the target interval represents a cheaper price range than the price of the "ideal" house in the catalog. We fixed the characteristics on which the buyer is not willing to make any concession, and the characteristics which it is not possible to change (as in the previous use case), making a restricted CF query on the remaining characteristics.

\begin{table}[H]
\centering
\resizebox{\linewidth}{!}{
\begin{tabular}[t]{>{}lllll}
\toprule
\textbf{ } & \textbf{OverallCond} & \textbf{Exterior2nd} & \textbf{OpenPorchSF} & \textbf{MoSold}\\
\midrule
\cellcolor{gray!6}{\textbf{initial}} & \cellcolor{gray!6}{6} & \cellcolor{gray!6}{HdBoard} & \cellcolor{gray!6}{75.5} & \cellcolor{gray!6}{5}\\
\textbf{CF} & 5 & CmentBd & 72 & 8\\
\bottomrule
\end{tabular}}
\caption{\label{tab:buyer1} \small{Point of view of the buyer: minimal concessions to make to find a house whose target price is inside the target interval [\$ 350000, \$ 376669], and which is similar to a "ideal" house in the catalog costing \$ 414419. The buyers are only ready to make concessions on the rating of finish and quality of materials, and on the outside configuration of the house. The CF approach suggests that the buyer should make some concessions on the material used for the second floor exterior covering (Exterior2nd), on the area of the open porch (OpenPorchSF), on the month when the acquisition should be made (MoSold), and on the overall condition of the house (OverallCond). Budget estimation: \$ 376420.}}
\end{table}

\begin{table}[H]
\centering
\resizebox{\linewidth}{!}{
\begin{tabular}[t]{>{}llllll}
\toprule
\textbf{ } & \textbf{OverallCond} & \textbf{BsmtFinType1} & \textbf{BsmtFinSF1} & \textbf{Functional} & \textbf{WoodDeckSF}\\
\midrule
\cellcolor{gray!6}{\textbf{initial}} & \cellcolor{gray!6}{5} & \cellcolor{gray!6}{GLQ} & \cellcolor{gray!6}{1085} & \cellcolor{gray!6}{Typ} & \cellcolor{gray!6}{192}\\
\textbf{CF} & 4 & LwQ & 915.44 & Sev & 142.99\\
\bottomrule
\end{tabular}}
\caption{\label{tab:buyer2} \small{Point of view of the buyer: minimal concessions to make to find a house whose target price is inside the target interval [\$ 200000, \$ 235247], and which is similar to a "ideal" house in the catalog costing \$ 257897. The buyers are only ready to make concessions on the house condition and on the non-living areas. The CF approach suggests that the buyer should make some concessions on the overall house condition (OverallCond), on the basement superficy (BsmtFinSF1) and finished area rating (BsmtFinType1), and on the home functionality (Functional) ... Budget estimation: \$235204.}}
\end{table}

\subsubsection{Case 3} \label{sec:use_case_3_3}

We use the possibility provided by the method to make explicit the decision regions of the model. Mathematically, the problem solved is to find all the regions $B_i^D$ (see notations in sec.~\ref{sec:decomposition}) such that:
\begin{align} \label{eq:domain_query}
\exists Y \in B_i^D \textnormal{ such that }
\begin{cases}
F(Y) \in \left[ F(X) - \epsilon, F(X) + \epsilon \right] \\
\norm{Y - X}_2^2 < \nu
\end{cases}
\end{align}
where $F$ is the regression model, and $\nu$ an upper bound on the maximal distance of the retrieved boxes $B_i^D$ to a query point $X$. Concretely, it amounts to make explicit the decision regions of the models (under the form of multi-dimensional intervals) for which the model's prediction is close to $F(X)$. We restrict the search around $X$ since there can be billions of such regions for large models.

\begin{figure}[H]
\centering
\includegraphics[width=0.35\textwidth]{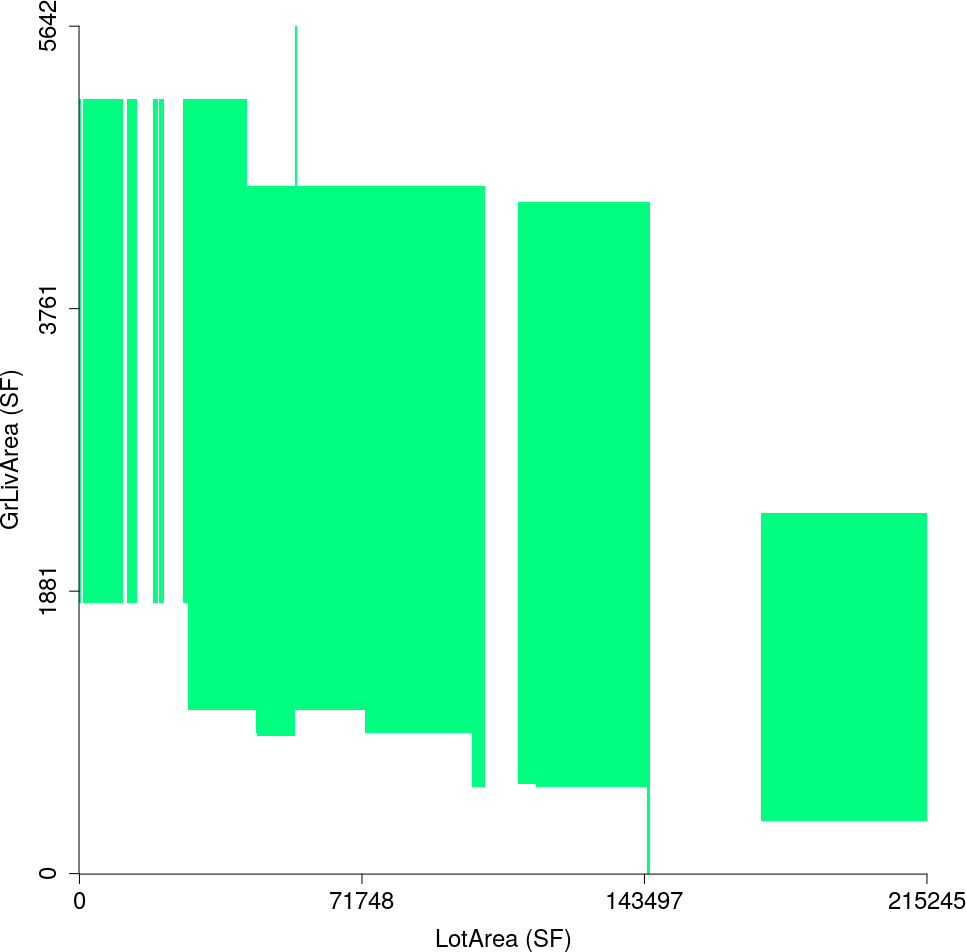}
\caption{\label{fig:decision_space_2D} \small{Projection on two characteristics ("GrLivArea" and "LotArea") of the decision regions where the model predicts a target price of \$ 200000.}}
\end{figure}

The fig.~\ref{fig:decision_space_2D} represents the projection on two characteristics ("GrLivArea" and "LotArea") of the decision regions where the model predicts a target price of \$ 200000. The diagram globally shows that, for a fixed price of \$ 200000, if a buyer wants a bigger lot area, he has to make concessions on the ground living area, and vice versa.

\subsection{Discussion}
In the provided scenarios, all the input characteristics are treated equally in the CF example computation. The user is only given the possibility to fix the value of some characteristics. We could include more prior knowledge about the problem at hand by adding the possibility to down/up-weight input characteristics, accounting that way for the fact that some characteristics are less/more important than others. This could be done by changing the CF problem to:
\begin{align} \label{eq:request_with_weights}
\textnormal{CF}(X, j) = \arg \min_{\left\{Y \left| \textnormal{Class}(F(Y)) = j \right.\right\}} (X-Y)^T D_\alpha (X-Y)
\end{align}
where $D_\alpha$ is a diagonal matrix containing the importance weights $\alpha$ associated with each characteristic. Solving this problem does not require any particular modification of the algo.~\ref{algo:request}: we would just need to change the instruction line 17 to: $dist_i \leftarrow dist_{cur} + \alpha[d_{cur}] \; dist\left( X[d_{cur}], \left[ bs_i, be_i \right] \right)$. This functionality is provided in the implementation.

\section{Conclusion and perspectives}

In this paper, we described a method for explicitly and exactly computing the decision regions of a tree-ensemble model. The method provides a simple geometric characterization of the domain of such a model in terms of pure decision regions, i.e. regions over which the model makes a constant prediction over all classes. From this characterization, we show that it is possible for a given query point to find the closest "virtual" point in terms of Euclidean distance that is classified in a user-specified class (different from the class of the query point) by the model. Using CF reasoning, it is then possible to produce diagnoses of faulty data in fault detection scenario by looking at what would have had to be changed a minima in the input data so that it be classified as "normal".

Both algorithmic and implementation optimizations render the CF example search problem tractable for arbitrarily large  tree-ensemble models, allowing the deployment on real life industrial fault detection problems.

In a future part of this work, we plan to extend the approach to model debugging. In the fault detection scenario, we indeed make the assumption that an abnormal detection/classification made by the model comes from an abnormal measurement in the input data. For model debugging, we would take the other point of view that an abnormal classification is a model failure to classify the data in the right class. Then, since our approach is making the decision regions of the model explicit, we could locate/characterize the regions of the input space where the model is making wrong decisions, and, either correct the model, or "correct" the training data that caused such wrong decision regions to appear. This last approach to model debugging would be called "dataset debugging", and would amount to search for natural adversarial examples in the training data that caused the model to learn erroneous decision regions.

\appendices

\section{Algorithmic proofs}

In this appendix, we provide the proofs relative to the recursive construction (dimension by dimension) of the search tree introduced in the fig.~\ref{fig:intersect_all}. Throughout the appendix, we use the notations introduced in sec.~\ref{sec:decomposition}.

\begin{figure}
\centering
\includegraphics[width=0.4\textwidth]{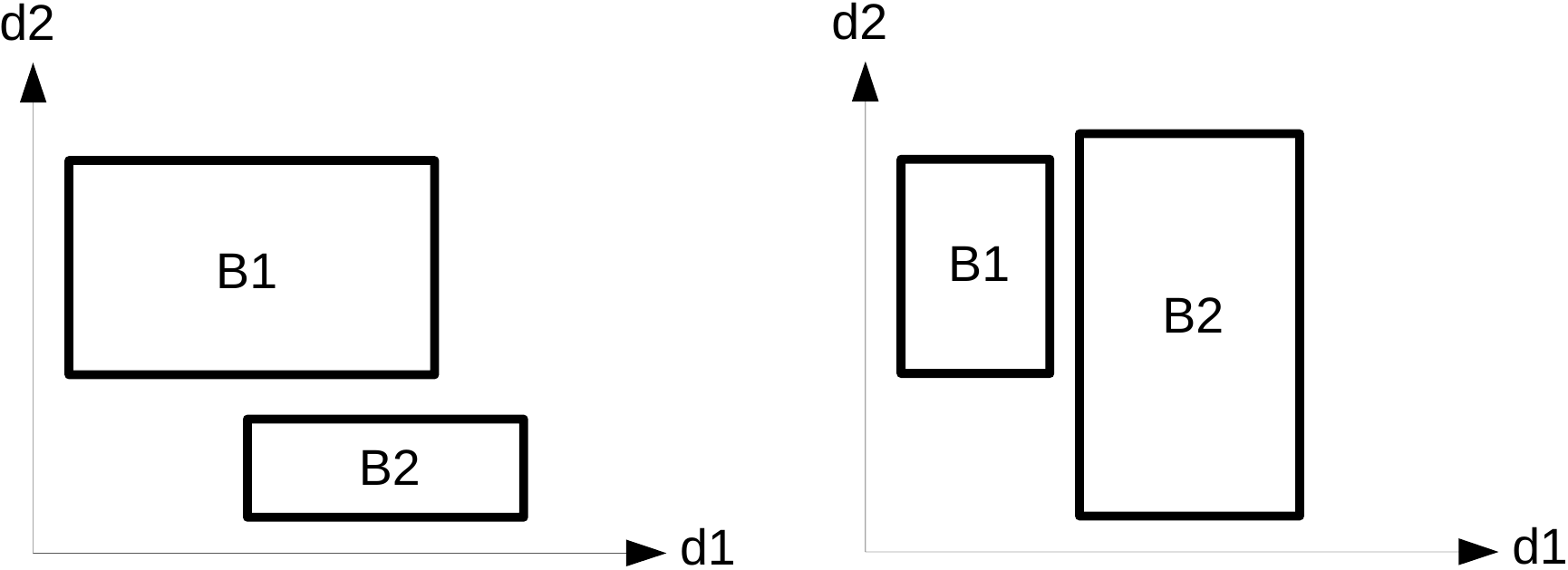}
\caption{\label{fig:no_intersect} \small{Illustration of prop.~\ref{prop:no_intersect}: two boxes not intersecting in one dimension have a null intersection (even if they intersect in the other dimensions. On the left diagram, B1 and B2 intersect in dimension 1 but not in dimension 2. On the right diagram, B1 and B2 intersect in dimension 2 but not in dimension 1. In both cases, the intersection of B1 and B2 is null.}}
\end{figure}

To add a new dimension in the search tree, we make use of the following property:
\theoremstyle{Proposition}
\begin{proposition} \label{prop:no_intersect}
if two boxes $B_i^D$ and $B_j^D$ do not intersect along a dimension $d$, $1 \leq d \leq D$, i.e. if $[bs_i^d, be_i^d] \cap [bs_j^d, be_j^d] = \emptyset$, then $B_i^D \cap B_j^D = \emptyset$.
\end{proposition}
This geometric property follows directly from the fact that, for any dimension $d$, a box $B_l^D$ is located between the two hyperplanes perpendicular to $\bf{u_d}$, the unit vector associated with the $d$-th coordinate axis, and passing one through the point $bs_l^d \; \bf{u_d}$, and the other through the point $be_l^d \; \bf{u_d}$.
The fig.~\ref{fig:no_intersect} illustrates the result of prop.~\ref{prop:no_intersect} in a two-dimensional feature space. 

\subsection{Proof of proposition \ref{prop:is_inter_max}}

\noindent\textbf{Proposition \ref{prop:is_inter_max}}: 
\textit{The boxes $B_{i, k}^{d+1}$ are maximum intersection boxes associated with the restriction of $F$ to the first $d+1$ dimensions.}

\noindent\textbf{Proof:} \newline
It is obvious that the sets $B_{i,k}^{d+1}$ are boxes in the sense of def.~\ref{eq:boites}. We then notice that each interval $\left[ bs_{i,q}^{d+1}, be_{i,q}^{d+1} \right]$ is a maximal intersection interval by construction. We then prove with the help of the prop.~\ref{prop:no_intersect} and the recurrence hypothesis made on $B_i^d$ ("$B_i^d$ is a maximal intersection box associated to $Restrict_{1\rightarrow d}(F)$") that the boxes $B_{i, k}^{d+1}$ are maximal intersection boxes associated with $Restrict_{1\rightarrow d+1}(F)$:
\begin{itemize}
    \item We show that: $B_{i, k}^{d+1} \subset \bigcap_{j \in J_{i, k}^{d+1}} F_j^{d+1}$ where $J_{i, k}^{d+1} = I_i^d \left( I_{i, k}^{d+1} \right)$:
    \begin{itemize}
        \item According to the recurrence hypothesis, $B_i^d \subset \bigcap_{j \in I_i^d} F_j^d$, so, $\forall j \in  I_i^d, B_i^d \subset F_j^d$
        \newline $\implies \; \forall j \in J_{i, k}^{d+1}, \; B_{i, k}^{d+1} \subset \left\{ F_j^d, \left[ bs_{i,k}^{d+1}, be_{i,k}^{d+1} \right] \right\}$, since, by  construction, $J_{i, k}^{d+1} \subset I_i^d$.
        \item By construction, $\forall j \in J_{i, k}^{d+1}, \; \left[ bs_{i,k}^{d+1}, be_{i,k}^{d+1} \right] \subset \left[ ls_j^{d+1}, le_j^{d+1} \right]$, so, $\forall j \in J_{i, k}^{d+1}, \;  B_{i, k}^{d+1} \subset F_j^{d+1} =\left\{ F_j^d, \left[ ls_j^{d+1}, le_j^{d+1} \right] \right\}$, which proves that $B_{i, k}^{d+1} \subset \bigcap_{j \in J_{i, k}^{d+1}} F_j^{d+1}$.
    \end{itemize}
    \item We show that: $\forall l \notin I_{i, k}^{d+1}$, $F_l^{d+1} \cap B_{i, k}^{d+1} = \emptyset$:
    \begin{itemize}
        \item Case $l \notin I_i^d$: $F_l^d \cap B_i^d = \emptyset$ according to the recurrence hypothesis. So, by using the prop.~\ref{prop:no_intersect}, we obtain that $F_l^{d+1} \cap B_{i, k}^{d+1} = \emptyset$.
        \item Case $l \in I_i^d \; \backslash \; J_{i, k}^{d+1}$: by construction, $\left[ ls_l^{d+1}, le_l^{d+1} \right] \cap \left[ bs_{i,k}^{d+1}, be_{i,k}^{d+1} \right] = \emptyset$, so, by using the fact that $F_l^{d+1} =\left\{ F_l^d, \left[ ls_l^{d+1}, le_l^{d+1} \right] \right\}$ and the prop.~\ref{prop:no_intersect}, we obtain that $F_l^{d+1} \cap B_{i, k}^{d+1} = \emptyset$.
    \end{itemize}
\end{itemize}

\subsection{Proof of proposition \ref{prop:rec_relation}}

\noindent\textbf{Proposition \ref{prop:rec_relation}:}
\textit{The decomposition $\left\{ \left\{ B_{i, k}^{d+1} \right\}_{k \in 1, \ldots, N_i^{d+1}} \right\}_{i \in 1, \ldots, L_d}$ obtained from all maximum intersection boxes $B_i^d$ is a decomposition into maximum intersection boxes of the leaves restricted to the first $d+1$ dimensions.}

\noindent\textbf{Proof:} \newline
\begin{itemize}
    \item The sets $B_{i, k}^{d+1}$ are maximal intersection boxes associated with $Restrict_{1\rightarrow d+1}(F)$ (cf. prop.~\ref{prop:is_inter_max}).
    \item We prove that: $\forall (i,j) \in \{1, \ldots, L_d\}$ et $\forall (q, l) \in \{1, \ldots, N_i^{d+1}\}$ tq. $q \neq l$, $B_{i, q}^{d+1} \cap B_{j, l}^{d+1} = \emptyset$:
    \begin{itemize}
        \item Case $i == j$: $\left[ bs_{i,q}^{d+1}, be_{i,q}^{d+1} \right] \cap \left[ bs_{i,l}^{d+1}, be_{i,l}^{d+1} \right] = \emptyset$ by construction, so, according to the prop.~\ref{prop:no_intersect}, $B_{i, q}^{d+1} \cap B_{i, l}^{d+1} = \emptyset$.
        \item Case $i \neq j$: $B_i^d \cap B_j^d = \emptyset$ (recurrence hypothesis), so, according to the prop.~\ref{prop:no_intersect}, and, by construction of $B_{i, q}^{d+1}$, $B_{i, q}^{d+1} \cap B_{j, l}^{d+1} = \emptyset$.
    \end{itemize}

    \item We prove that: $\bigcup_{i = 1, \ldots, L_d} \bigcup_{k = 1, \ldots, N_i^{d+1}} B_{i, k}^{d+1} = Restrict_{1\rightarrow d+1}(F)$:
    \begin{itemize}
    \item Given the fact that the boxes $B_{i, k}^{d+1}$ are maximal intersection boxes associated with $Restrict_{1\rightarrow d+1}(F)$, we have: $\bigcup_{i = 1, \ldots, L_d} \bigcup_{k = 1, \ldots, N_i^{d+1}} B_{i, k}^{d+1} \subset Restrict_{1\rightarrow d+1}(F)$
    \item We then show that $\forall X \in Restrict_{1\rightarrow d+1}(F)$, $\exists l \in \{1, \ldots, L_d\} \; \exists q \in \{1, \ldots, N_l^{d+1}\} \textnormal{ such that } X \in B_{l, q}^{d+1}$:
    \begin{itemize}
        \item $\bigcup_{i \in 1, \ldots, L_d} I_i^d = \left\{ 1, \ldots, N \right\}$ and $\forall i \in \{1, \ldots, L_d\}, \; \bigcup_{n \in I_i^d} \left[ ls_n^{d+1}, le_n^{d+1} \right] = \bigcup_{q \in 1, \ldots, N_i^{d+1}} \left[ bs_{i,q}^{d+1}, be_{i,q}^{d+1} \right]$, so, $\bigcup_{i \in 1, \ldots, L_d} \bigcup_{k \in 1, \ldots, N_i^{d+1}} \left[ bs_{i,k}^{d+1}, be_{i,k}^{d+1} \right] = \bigcup_{m \in 1, \ldots, N} \left[ ls_m^{d+1}, le_m^{d+1} \right]$
        \newline $\implies \exists ! (l, q) \textnormal{ such that } X[d+1] \in \left[ bs_{l,q}^{d+1}, be_{l,q}^{d+1} \right]$
        \item $\bigcup_{i \in 1, \ldots, L_d} B_i^d = Restrict_{1\rightarrow d}(F) = \bigcup_{j \in 1, \ldots, N} F_j^d$ according to the recurrence hypothesis
        \newline $\implies \exists p \in \{1, \ldots, L_d\} \textnormal{ such that } X[1:d] \in B_p^d$. 
        \newline Moreover, we necessarily have $p=l$:
        \begin{itemize}
            \item $X[1:d] \in B_p^d \implies X \in \bigcap_{i \in I_p^d} F_j^{d+1}$
            \item $X[d+1] \in \left[ bs_{l,q}^{d+1}, be_{l,q}^{d+1} \right] \implies X \notin \bigcup_{n \in \{1,\ldots, N\} \; \backslash \; J_{l, q}^{d+1}} F_n^{d+1}$
            \item Yet, $\forall l \neq p, \; \bigcap_{i \in I_p^d} F_j^{d+1} \subset \bigcup_{n \in \{1,\ldots, N\} \; \backslash \; J_{l, q}^{d+1}} F_n^{d+1} $ since $I_p^d \subset \{1,\ldots, N\} \; \backslash \; J_{l, q}^{d+1}$
            \newline $\implies p=l$
        \end{itemize}
        We then have $X \in B_{p, q}^{d+1}$, which proves the inclusion:  $Restrict_{1\rightarrow d+1}(F) \subset \bigcup_{i = 1, \ldots, L_d} \bigcup_{k = 1, \ldots, N_i^{d+1}} B_{i, k}^{d+1}$, and, finally, that: $\bigcup_{i = 1, \ldots, L_d} \bigcup_{k = 1, \ldots, N_i^{d+1}} B_{i, k}^{d+1} = Restrict_{1\rightarrow d+1}(F)$.
    \end{itemize}
    \end{itemize}
\end{itemize}

\bibliographystyle{IEEEtran}
\bibliography{./bibtex/my_bib}

\end{document}